\documentclass[10pt,twocolumn,letterpaper]{article}

\usepackage[pagenumbers]{cvpr}      %

\usepackage{colortbl}
\usepackage{placeins}
\usepackage[normalem]{ulem}

\definecolor{wcTableBest}{HTML}{D8ECFA}
\definecolor{wcTableSecond}{HTML}{EAF5FC}
\definecolor{wcTableThird}{HTML}{F4F9FD}
\newcommand{\wcBest}[1]{\cellcolor{wcTableBest}\textbf{#1}}
\newcommand{\wcSecond}[1]{\cellcolor{wcTableSecond}#1}
\newcommand{\wcThird}[1]{\cellcolor{wcTableThird}#1}
\newcommand{\wcTableFont}{\small}

\newcommand{\wbhu}[1]{#1}

\usepackage{algorithm}
\usepackage{algpseudocode}

\patchcmd{\abstract}
  {\centerline{\large\bf Abstract}}
  {\centerline{\large\bf\raisebox{-6pt}[\height][\depth]{Abstract}}}
  {}{\PackageError{WorldCrafter}{Could not adjust Abstract heading}{}}

\definecolor{cvprblue}{rgb}{0.21,0.49,0.74}
\usepackage[pagebackref,breaklinks,colorlinks,allcolors=cvprblue]{hyperref}

\def\confName{CVPR}
\def\confYear{2027}

\title{WorldCrafter: Consistent Video World Model with Implicit 3D-aware Memory}

\author{
Wangbo Yu$^{1*}$\hspace{0.9em}%
Kunhao Liu$^{1*}$\hspace{0.9em}%
Wenbo Hu$^{1\dagger}$\hspace{0.9em}%
Shenghai Yuan$^{2}$\hspace{0.9em}%
Chaoran Feng$^{2}$\hspace{0.9em}%
Haiyang Zhou$^{2}$\\
Yukun Huang$^{1}$\hspace{0.9em}%
Yiran Wang$^{1}$\hspace{0.9em}%
Wang Zhao$^{1}$\hspace{0.9em}%
Yingmin Luo$^{1}$\hspace{0.9em}%
Ying Shan$^{1}$\\[0.4em]
$^{1}$ARC Lab, Tencent IEG \qquad $^{2}$Peking University\\[0.3em]
{\large\url{https://drexubery.github.io/WorldCrafter}}
}

\begin{document}
\twocolumn[{
    \renewcommand\twocolumn[1][]{##1}
    \maketitle
    \vspace{-2.5em}
    \centering
    \includegraphics[width=\textwidth]{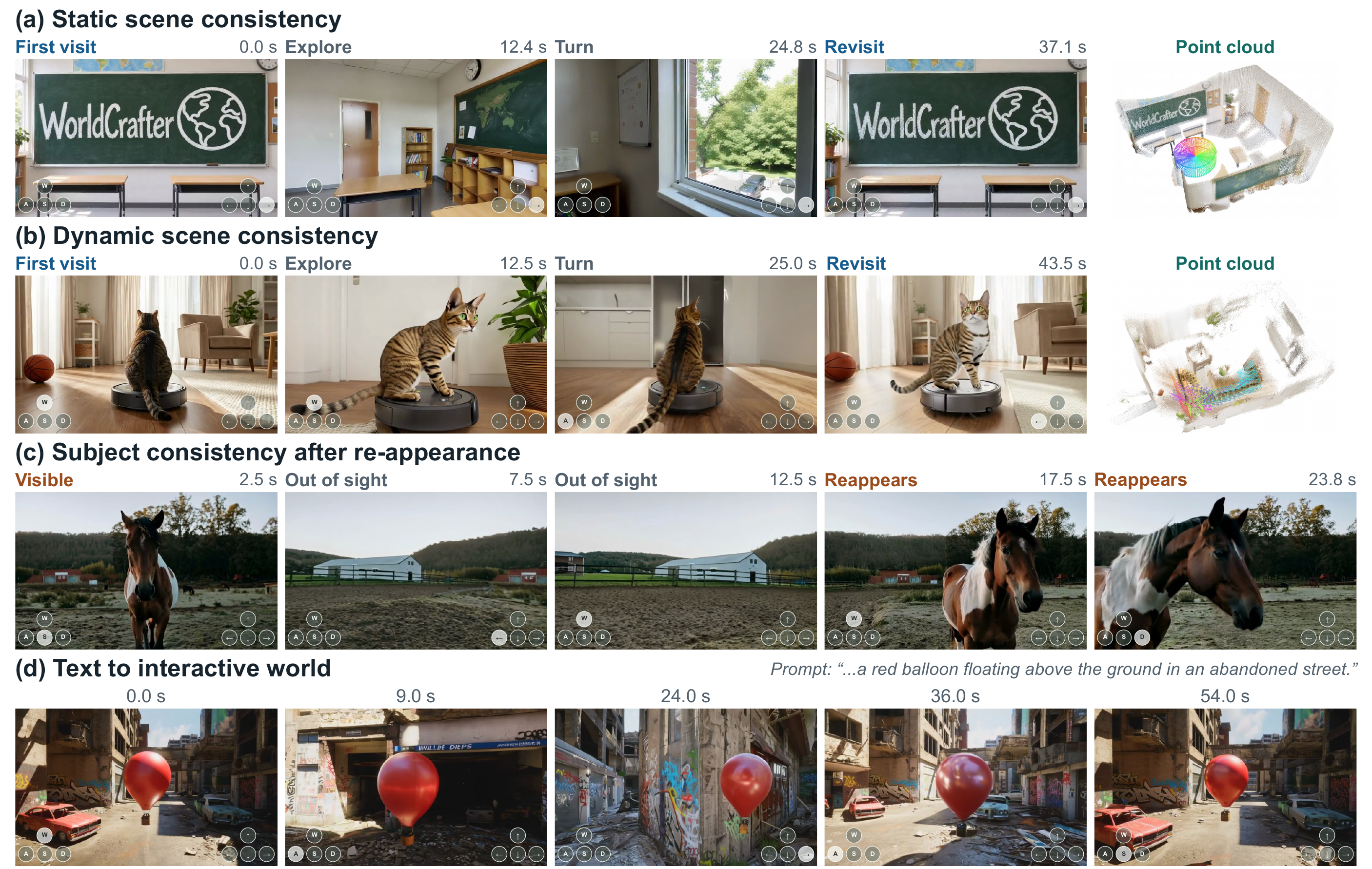}
    \vspace{-1.5em}
    \captionof{figure}{
        \wbhu{WorldCrafter enables consistent, camera-controlled exploration. It preserves scene appearance and structure across revisits in static (a) and dynamic (b) scenes, as illustrated by reconstructed point clouds. Subjects remain consistent after leaving and re-entering view (c). Users can also generate and explore scenes from text descriptions alone (d).}
        }
    \label{fig:teaser}
    \vspace{1em}
}]

\iftoggle{cvprfinal}{%
\begingroup
\renewcommand{\thefootnote}{\fnsymbol{footnote}}
\begin{NoHyper}
\footnotetext[1]{Equal contribution.\qquad\textsuperscript{\textdagger}Corresponding author.}
\end{NoHyper}
\endgroup
}{}
\addtocontents{toc}{\protect\setcounter{tocdepth}{-2}}
\begin{abstract}
\wbhu{Video world models enable interactive exploration of dynamic environments, yet struggle to respect prior observations over long horizons and across viewpoints.}
We present \emph{WorldCrafter}, a video world model that learns a camera-queryable implicit 3D-aware memory for this purpose.
The key insight is to let the requested viewpoint shape how multi-view evidence is compressed into the video generator's limited token budget.
Trained jointly with the video generator, a memory encoder and pose-conditioned readout module integrate historical observations into a fixed set of target view-specific tokens before denoising, without explicit depth-based correspondences.
By combining this memory with recent temporal context and few-step distillation, WorldCrafter enables streaming scene exploration from a single input image or text prompt.
\wbhu{Experiments across static and dynamic scenes show substantial gains in long-horizon consistency and camera-control accuracy while preserving visual quality during minute-scale exploration.}
\end{abstract}

\section{Introduction}
\label{sec:intro}

Video world models enable interactive exploration of dynamic environments by generating new observations as users move the camera~\cite{genie3,hyworld2025,happyoyster,oasis,parkerholder2024genie2,li2025hunyuangamecraft,matrixgame3,mao2025yume,zhu2026sanawm}.
\wbhu{Maintaining a coherent world requires memory beyond the recent context so that previously observed content remains consistent when revisited, even from a different viewpoint.}

\wbhu{A straightforward way to provide memory for video world models is to include previously generated frames in attention.}
Full-history attention incurs substantial computation costs~\cite{hong2025relic,cai2025mixturecontexts,yi2026worldkv}, while FoV-based \wbhu{ history retrieval trades view coverage for efficiency~\cite{xiao2025worldmem,yu2025context,yi2026worldkv,cai2025mixturecontexts,chen2026out,li2025vmem,lyra2026,wu2026infinite,matrixgame3,sun2025worldplay}}.
Explicit spatial memories provide a shared 3D reference but depend on accurate geometry and struggle with dynamic scenes~\mbox{\cite{wu2025spatialmemory,zhao2025spatia,yin2026evoke,yu2026mosaicmem}}. \wbhu{Implicit memories compress history into learned representations~\cite{savov2025statespacediffuser,yu2025videossm}}, with recent methods incorporating geometry features for 3D awareness~\cite{vggt,wei2026gim,huang2026cinescene}. However, these geometry-oriented representations prioritize geometric prediction over the appearance fidelity needed to reproduce previously observed scenes~\cite{szymanowicz2026lagernvs}.

Recent advances in 3D representation learning~\cite{szymanowicz2026lagernvs,jiang2025rayzer,kim2026svsm,jin2025lvsm} have demonstrated remarkable capabilities in learning compact scene representations, making them a natural foundation for the memory space of video world models. Motivated by this, we present \textbf{WorldCrafter}, a video world model with implicit 3D-aware memory for consistent interactive generation. At its core, a memory encoder initialized from pretrained 3D representation encoders~\cite{szymanowicz2026lagernvs,kim2026svsm,kaashyap2026less} maps historical latent frames into a compact memory space, inheriting their learned 3D inductive bias. We further optimize this memory space by jointly training the memory encoder, video diffusion transformer (DiT), and \wbhu{a memory readout module, enabling the memory to co-adapt with the DiT token space.}

\wbhu{To extract generation-relevant information from this memory, we compare pose-free and pose-guided readout under a fixed token budget.
Pose-guided readout focuses on information relevant to the requested viewpoints and yields better revisit consistency and camera control in our experiments.
The resulting tokens condition the DiT directly through self-attention, without reconstructing target-view images.}
During interaction, we exploit a max-coverage strategy to select complementary historical views according to their joint camera coverage and combine the queried memory with recent temporal context.
The memory supplies historical scene information, while recent context supports the continuation of visible motion.
\wbhu{Generated chunks are added to the history archive, while the encoder input size and DiT memory-token budget remain fixed.}
Together with camera-conditioned autoregressive generation and few-step distillation, our model achieves real-time streaming inference  while preserving precise camera control and minute-scale consistency under complex user-specified camera trajectories.

Our contributions are as follows:
\begin{itemize}
    \item We introduce an implicit 3D-aware memory mechanism for video world models. It learns to encode history latent frames into a compact memory representation that preserves spatial-temporal context, enabling efficient memory writing and readout within a fixed token budget.
    \item
    \wbhu{We integrate this memory mechanism into camera-controllable autoregressive video generation, achieving leading revisit consistency (47.6\% improvement relative to the strongest baseline) and camera-control accuracy, with controlled ablations supporting our design.}
    \item We build a real-time interactive system through few-step
distillation, achieving streaming inference  while maintaining visual quality
throughout minute-scale exploration.

\end{itemize}

\begin{figure*}[t]
    \centering
    \includegraphics[width=\textwidth]{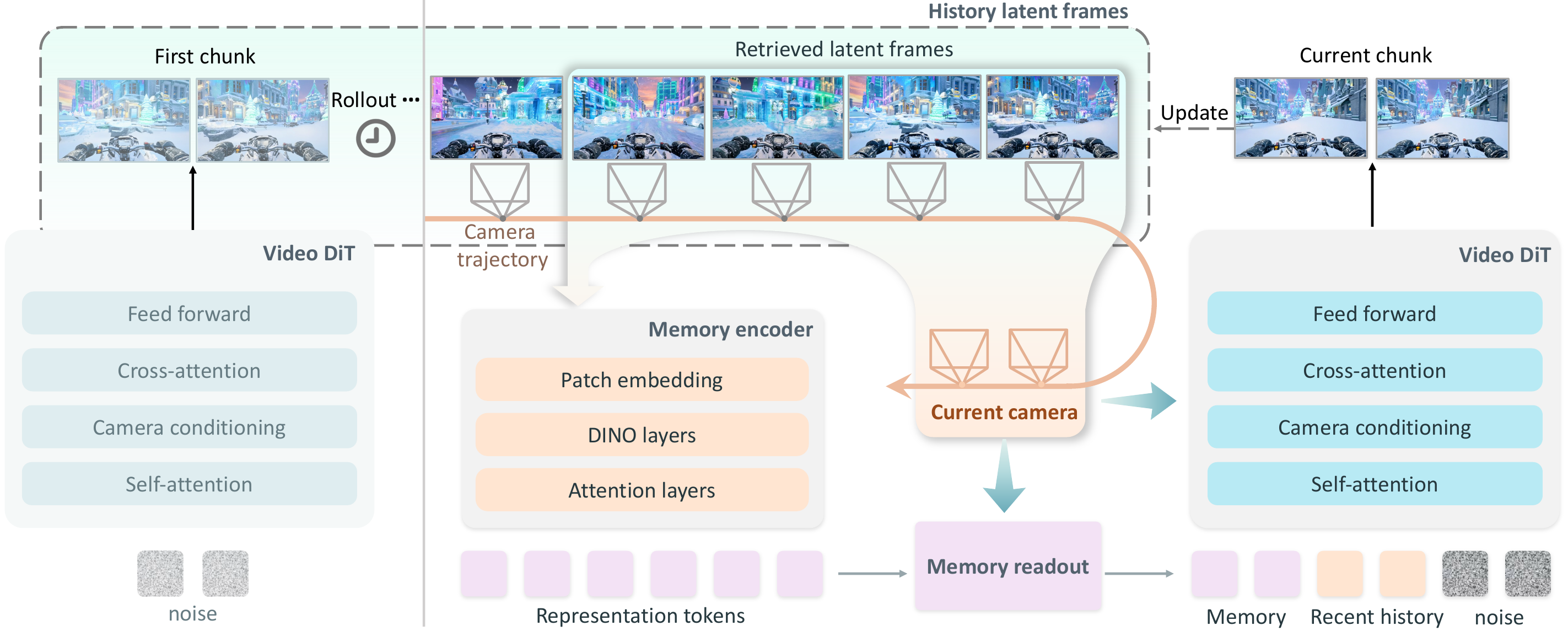}
    \vspace{-1em}
    \caption{\textbf{Overview of the WorldCrafter pipeline.} The video DiT generates the first chunk conditioned on camera poses, before any history is available. At subsequent rollout steps, max-coverage history retrieval selects latent frames based on the current camera poses, and the memory encoder maps them into a compact 3D-aware representation. A pose-conditioned memory readout module extracts fixed-size memory tokens that, together with recent history, guide generation of the current chunk. Generated frames are added to the history for subsequent rollouts.}
    \label{fig:pipeline}
    \vspace{-1em}
\end{figure*}
\section{Related Work}
\label{sec:related_work}

\subsection{Interactive Video World Models}

Video world models generate future observations in response to user actions, turning video generation into interactive simulation\wbhu{~\cite{huang2025towards,bruce2024genie,alonso2024diffusion,valevski2024diffusion,che2024gamegen,zhang2026matrixgame4,zhang2026playgen2,savva2026solaris,po2026multigen,pondaven2026actionparty,tong2026scope,hu2026mira,hu2026wantofight,lin2026stateplay,cai2026mass,ye2026worldcast,li2026overseer,xu2026consistworld,dai2025fantasyworld}}. Recent methods increasingly combine streaming generation with camera control to support real-time interaction and long-horizon exploration~\cite{genie3,happyoyster,gao2026lingbotworld2,zhu2026sanawm,matrixgame3,matrixgame35,team2026dreamxworld,jiang2026abotworld,sun2025worldplay,lyra2026,yin2026evoke,alayaworld2026,mao2025yume,nvidia2026cosmos3,chen2026reworld,zhang2026echowm,huang2026solarwmopendatascalable,feng2024matrix,he2025matrixgame2,team2026lingbotworld,zhu2025astra,xu2026wonder,nam2026worldcam,chou2026captainsafari,li2026magicworld,zhang2026worldplay2,rajpal2026waypoint,chen2026hynomas}.

\wbhu{Streaming generation extends video diffusion through rolling denoising or temporally varying noise levels~\cite{kim2024fifodiffusion,chen2024diffusionforcing,ruhe2024rolling,teng2025magi,chen2025skyreelsv2}, while autoregressive distillation and rollout-aware training improve sampling efficiency and mitigate error accumulation~\cite{yin2025causvid,huang2025selfforcing,liu2025rollingforcing,zhu2026causalforcing,zhao2026causalforcingpp,li2026forgewm}. Flexible history conditioning supports longer rollouts~\cite{song2025historyguided,henschel2025streamingt2v,ji2025memflow,hu2026longliverag}, complemented by efficient streaming designs and parallel implementations~\cite{yang2026longlive,chen2026longlive2,zhao2026minwm,liu2026worldpager}. }

Camera control is commonly implemented through discrete action inputs\wbhu{~\cite{oasis,yu2025gamefactory,li2025hunyuangamecraft,matrixgame3,mao2025yume15,jiang2026abotworld}}, continuous camera parameters\wbhu{~\cite{wang2024motionctrl,he2024cameractrl,bai2025recammaster,li2025prope,zhang2025ucpe,bahmani2025ac3d,xu2024camco,bahmani2025vd3d,he2025cameractrl2,xiang2026viewrope,fu2026plenoptic}}, or point-cloud renders along the target trajectory~\cite{yu2024viewcrafter,ren2025gen3c,yu2025trajectorycrafter}. Although these signals specify viewpoint changes, they do not provide a persistent state that preserves scene content beyond the context window. Thus, combining streaming generation with camera control alone remains insufficient for consistent long-horizon exploration.

\subsection{Memory Mechanisms in Video World Models}

Persistent video world models require memory beyond the recent video context~\cite{king2026echomemory}. We categorize existing approaches according to their stored representations: context memory, spatial memory, and implicit memory.

Context memory retains historical frames\wbhu{, latent tokens, or cached attention features} for attention-based reuse\wbhu{~\cite{xiao2025worldmem,yu2025context,matrixgame3,sun2025worldplay,zhang2025framepack,oshima2025worldpack,hong2025relic,cai2025mixturecontexts,wu2026worldtrace,chen2026reworld,chen2025vrag,yang2026decmem,guo2026memorize,yang2026gear,xu2026glimpse,dou2026slotmemory,wang2026redate}}.
Frame-retrieval methods select a small subset using camera overlap or reconstructed surfaces\wbhu{~\cite{yu2025context,li2025vmem,huang2025memory,gao2026memcam,joo2026covrag,seo2026seoulworld,fu2026plenoptic,peng2026car}}.
\wbhu{These methods learn to access historical context within the generator. WorldCrafter instead aggregates history into a 3D-aware memory representation using a pretrained multi-view encoder, then reads out fixed-size memory tokens before denoising.}

Spatial memory instead transforms \textcolor{black}{historical observations into conditions aligned with target camera poses}~\cite{yu2024viewcrafter,ren2025gen3c,wu2025spatialmemory,zhao2025spatia,yin2026evoke,wang2026anchorweave,li2026i3dm,yu2026mosaicmem,xu2026ucm,huang2025voyager,chen2025deepverse,zhang2026worldstereo,wang2026latentspatialmemory,zhang2026fourlc,he2026ppmem,choi2026l3gworld,fang2026agiworld,lv2026octworld,wang2026evoworld,wan2026totem}. Representative methods perform this transformation via novel view synthesis, encode the synthesized target-view frames into the VAE latent space, and concatenate the resulting latents channel-wise with the input noise to condition video diffusion models~\cite{yu2024viewcrafter,ren2025gen3c,wang2026anchorweave,li2026i3dm,wu2025spatialmemory}. The resulting spatial correspondence across viewpoints facilitates revisiting previously observed regions. However, strong alignment to the target view can overconstrain scene dynamics, limiting their ability to model dynamic objects and environments.

Implicit memory encodes history into learned representations~\cite{savov2025statespacediffuser,yu2025videossm,wu2026infinite,chen2026out,li2026walkingimplicit,garcin2026persist,hu2026learningretrieve,shi2026honeycomb,wu2026worldt3}.
Existing methods update memory recurrently alongside local context\wbhu{~\cite{savov2025statespacediffuser,yu2025videossm,po2025longcontext,wu2025pack,lillemark2026flowequivariant,xia2026loci,yu2025malt,chen2026rad,li2026arl2,bian2026echoinfinity,lee2025edeline}} or compress observations using learned encoders~\cite{wu2026infinite,chen2026out,zhu2025memorizegenerate,zhang2026tinyhistory,yu2026memlearner}.
\wbhu{Geometry-aware approaches draw on pretrained geometry estimators such as VGGT~\cite{vggt,wu2026geometryforcing}: CineScene~\cite{huang2026cinescene} uses their features as generation conditions, while GIM-World~\cite{wei2026gim} distills them into memory through geometric supervision.}
\wbhu{However, geometry-estimation pretraining prioritizes geometric prediction over the appearance fidelity needed for consistent visual recall~\cite{szymanowicz2026lagernvs}.}
\wbhu{WorldCrafter instead adapts multi-view scene representations learned through novel-view reconstruction, which requires preserving both geometry and appearance.}
\wbhu{Initialized from LagerNVS~\cite{szymanowicz2026lagernvs}, our memory encoder and readout inherit a learned 3D inductive bias and are jointly optimized with the video generator.}

\section{Method}
\label{sec:method}

\subsection{Preliminary}
\label{sec:method_preliminary}

\noindent\textbf{Latent video diffusion.}
Let $\mathbf{x} \in \mathbb{R}^{3 \times F \times H \times W}$ denote a clean video of $F$ frames.
A video variational autoencoder (VAE)~\cite{kingma2014autoencoding} maps it to a spatiotemporal latent $\mathbf{z} = \mathcal{E}_{\mathrm{VAE}}(\mathbf{x}) \in \mathbb{R}^{c \times f \times h \times w}$.
Its decoder reconstructs the video as $\hat{\mathbf{x}} = \mathcal{D}_{\mathrm{VAE}}(\mathbf{z})$.
Operating in this latent space substantially reduces the token sequence processed by the Diffusion Transformer (DiT)~\cite{peebles2023scalable}-based denoiser, which patchifies $\mathbf{z}$ into video tokens and applies 3D self-attention.
In a conventional bidirectional video diffusion model such as Wan 2.1~\cite{wan2025wan}, every video token can attend to all other tokens in the clip during each denoising step.

The denoiser is trained with conditional flow matching~\cite{lipman2023flow}.
For diffusion time $t \sim \mathcal{U}(0,1)$ and Gaussian noise $\boldsymbol{\epsilon} \sim \mathcal{N}(\mathbf{0},\mathbf{I})$, the noisy latent and its target velocity are
\begin{equation}
    \mathbf{z}_t = (1-t)\mathbf{z} + t\boldsymbol{\epsilon},
    \qquad
    \mathbf{v}^{*}_t = \boldsymbol{\epsilon} - \mathbf{z}.
    \label{eq:flow_path}
\end{equation}
Given a text condition $\mathbf{y}$, the DiT $\mathbf{v}_{\theta}$ predicts this velocity by minimizing
\begin{equation}
    \mathcal{L}_{\mathrm{FM}}
    = \mathbb{E}_{\mathbf{z},t,\boldsymbol{\epsilon}}
    \left[
        \left\|
        \mathbf{v}_{\theta}(\mathbf{z}_t,t\mid\mathbf{y})
        - \mathbf{v}^{*}_t
        \right\|_2^2
    \right].
    \label{eq:flow_matching}
\end{equation}

\noindent\textbf{Chunk-wise autoregressive video generation.}
To extend generation beyond a fixed clip, autoregressive methods divide a long video into fixed-length chunks and generate them sequentially.
At a generic rollout step, we omit the rollout index from all quantities for clarity.
Let $\mathbf{z}$ denote the clean latent of the current chunk, $\mathbf{z}_t$ its state at diffusion time $t$, $\mathbf{z}^{\mathrm{h}}$ the accumulated clean history frames, and $\mathbf{z}^{\mathrm{r}}$ the fixed-length recent history frames retained from $\mathbf{z}^{\mathrm{h}}$.
Given the text condition $\mathbf{y}$, a standard chunk-wise autoregressive model evolves the current latent through the conditional flow
\begin{equation}
    \frac{\mathrm{d}\mathbf{z}_t}{\mathrm{d}t}
    = \mathbf{v}_{\theta}\!\left(
        \mathbf{z}_t,t
        \mid \mathbf{z}^{\mathrm{r}},\mathbf{y}
    \right).
    \label{eq:ar_flow}
\end{equation}
At each denoising step, the DiT processes the concatenated latent sequence $[\mathbf{z}^{\mathrm{r}};\mathbf{z}_t]$, where $[\,;\,]$ denotes concatenation along the token sequence.
After denoising, $\mathbf{z}$ is appended to $\mathbf{z}^{\mathrm{h}}$, and the sliding window of $\mathbf{z}^{\mathrm{r}}$ is updated with $\mathbf{z}$.

\subsection{Model Architecture}
\label{sec:base_architecture}

To enable long-horizon memory and user interaction, we extend the standard autoregressive flow in Eq.~\eqref{eq:ar_flow} by conditioning each chunk jointly on a memory $\mathbf{M}$ derived from the accumulated history $\mathbf{z}^{\mathrm{h}}$ and a target camera trajectory $\mathbf{C}$:
\begin{equation}
    \frac{\mathrm{d}\mathbf{z}_t}{\mathrm{d}t}
    = \mathbf{v}_{\theta}\!\left(
        \mathbf{z}_t,t
        \mid \mathbf{M},\mathbf{z}^{\mathrm{r}},\mathbf{C},\mathbf{y}
    \right).
    \label{eq:memory_camera_ar_flow}
\end{equation}

The overview of our pipeline is shown in Figure~\ref{fig:pipeline}. The video DiT generates the first chunk conditioned on camera poses alone, as no history is yet available. As history accumulates, we learn a memory encoder to map history latent frames into a compact 3D-aware representation, which is read out as a fixed-size memory $\mathbf{M}$ to condition subsequent generation.

\noindent\textbf{Memory conditioning.} At each denoising step, the DiT processes $[\mathbf{M};\mathbf{z}^{\mathrm{r}};\mathbf{z}_t]$ as a single latent sequence.
This adds a dedicated memory stream to the recent history while preserving chunk-level causality.
Unlike prior context-based memory methods~\cite{xiao2025worldmem,yu2025context,sun2025worldplay,matrixgame3} that instantiate $\mathbf{M}$ as a fixed-length context retrieved from the accumulated history $\mathbf{z}^{\mathrm{h}}$, we model $\mathbf{M}$ by mapping $\mathbf{z}^{\mathrm{h}}$ into a compact 3D-aware implicit memory representation.

\noindent\textbf{Camera conditioning.} The target trajectory $\mathbf{C}$ specifies a camera-to-world pose and camera intrinsics for each frame.
Following PRoPE~\cite{li2025prope}, we encode relative camera geometry as a positional transformation within self-attention.
We implement this conditioning using the parallel camera-attention branch of UCPE~\cite{zhang2025ucpe}, which adopts independent query, key, and value projections and adds its output to the original self-attention through a zero-initialized projection.
The camera branch is applied only to the noisy part $\mathbf{z}_t$ of the concatenated sequence; the memory $\mathbf{M}$ and recent context $\mathbf{z}^{\mathrm{r}}$ are processed without camera injection.

\subsection{Memory Encoder}
\label{sec:memory_encoder}

\noindent\textbf{Memory writing.}
As shown in Figure~\ref{fig:pipeline}, after generating the first chunk, the memory encoder $\Phi$ writes the accumulated history latents $\mathbf{z}^{\mathrm{h}}$ and their corresponding camera parameters $\mathbf{C}^{\mathrm{h}}$ into an implicit 3D-aware representation $\mathbf{R}$:
\begin{equation}
    \mathbf{R}
    = \Phi(\mathbf{z}^{\mathrm{h}},
           \mathbf{C}^{\mathrm{h}})
    \in \mathbb{R}^{|\mathbf{z}^{\mathrm{h}}|L \times d},
    \label{eq:memory_representation}
\end{equation}
where $|\mathbf{z}^{\mathrm{h}}|$ denotes the number of history latent frames.
The encoder produces $L$ tokens of dimension $d$ per history latent frame.

We initialize the encoder architecture and weights from the LagerNVS encoder~\cite{szymanowicz2026lagernvs}, discarding its shallow image-processing layers and adding a new patch embedding layer to map each latent frame directly into its representation space.
The history camera poses are expressed relative to the latest latent frame in $\mathbf{z}^{\mathrm{h}}$ and injected as camera tokens during encoding.
The resulting representation tokens $\mathbf{R}$ aggregate geometry and appearance information across the input history without materializing an explicit 3D reconstruction.

Since the length of $\mathbf{R}$ grows linearly with the number of input history frames $|\mathbf{z}^{\mathrm{h}}|$, to bound the cost of the encoder, we restrict its input to $k$ history latent frames.
At inference, we retain the latest latent frame in $\mathbf{z}^{\mathrm{h}}$ and greedily select $k-1$ complementary frames whose joint field of view (FoV) maximizes coverage of the target region along the upcoming camera trajectory.
We denote the selected history latents and their camera parameters by $\mathbf{z}^{\mathrm{s}}$ and $\mathbf{C}^{\mathrm{s}}$, respectively.
Prior context-based memory methods~\cite{xiao2025worldmem,yu2025context,matrixgame3,sun2025worldplay} retrieve only a few history frames by ranking their pairwise FoV similarity to target poses~\cite{yu2025context,matrixgame3}, resulting in limited coverage and sensitivity to individual selections.
In contrast, our memory encoder accommodates more history frames under a comparable budget, while max-coverage history retrieval yields broader coverage and greater robustness. \textcolor{black}{The detailed retrieval algorithm is provided in \hyperref[sec:supp-network]{Supplementary Sec.~\ref*{sec:supp-network}}.}

\noindent\textbf{Memory readout.}
The written representation $\mathbf{R}$ should then be read out as a fixed-size memory $\mathbf{M}$ that conditions the DiT. We compare two readout mechanisms under a fixed DiT token budget.

The first is pose-free readout, in which a learned readout module maps the complete representation into a fixed set of memory tokens compatible with the DiT input:
\begin{equation}
    \mathbf{M}
    = \operatorname{Readout}\!\left(
        \Phi(\mathbf{z}^{\mathrm{s}},\mathbf{C}^{\mathrm{s}})
      \right).
    \label{eq:representation_compression}
\end{equation}
This readout is independent of the upcoming camera trajectory, leaving the DiT attention to identify information relevant to the current generation.

The second is pose-guided readout, which queries the representation using a fixed-size set of query poses $\mathbf{C}^{\mathrm{q}} \subset \mathbf{C}$ sampled from the upcoming target camera trajectory:
\begin{equation}
    \mathbf{M}
    = \operatorname{Readout}\!\left(
        \Phi(\mathbf{z}^{\mathrm{s}},\mathbf{C}^{\mathrm{s}}),
        \mathbf{C}^{\mathrm{q}}
      \right).
    \label{eq:target_pose_query}
\end{equation}
In pose-guided readout, we initialize the readout module with the shallow decoder layers of LagerNVS~\cite{szymanowicz2026lagernvs} and add projection layers to map the output tokens directly into the DiT token space.

Empirically, we find that pose-guided readout outperforms pose-free readout and yields more accurate camera control. We attribute this gain to a more effective allocation of the fixed memory budget to target-relevant information. We therefore adopt pose-guided readout, with ablations presented in Sec.~\ref{sec:ablation}.

\begin{figure*}[!t]
    \centering
    \includegraphics[width=.95\textwidth]{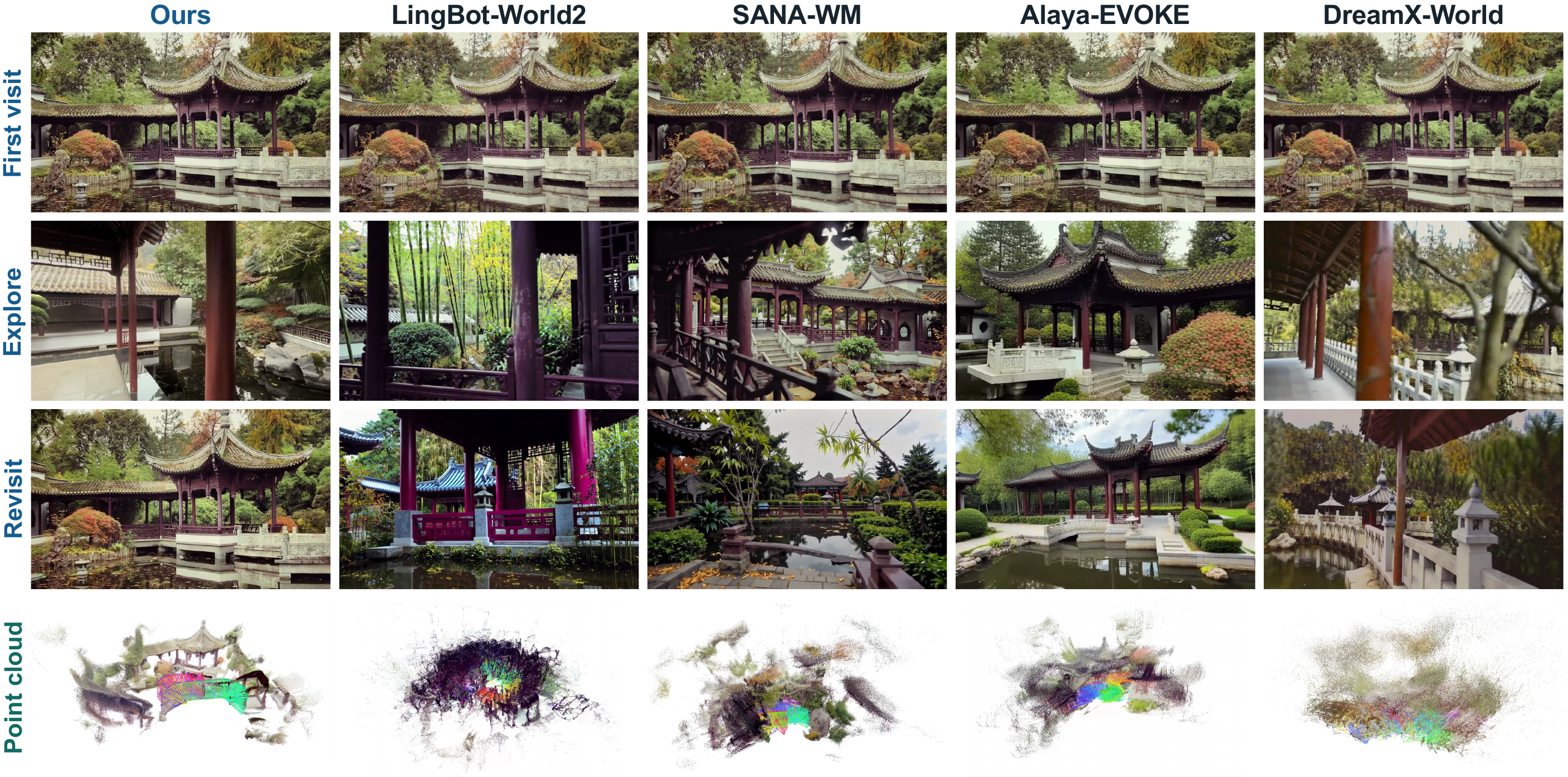}
    \caption{\textbf{Qualitative long-horizon revisit comparison in a static scene.} Rows show the first observation, an intermediate view during exploration, the matched revisit, and the point cloud reconstruction of the generated video, respectively.}
    \label{fig:revisit_qualitative}
    \vspace{0.5em}
    \raggedright
    \begin{minipage}[t]{0.49\textwidth}
        \captionsetup{type=table}
        \centering
  \caption{\textbf{Long-horizon revisit consistency.} Results are averaged over 725 generated videos using matched first-visit and revisit frames. Progressively lighter shades of blue mark the best, second-best, and third-best results, respectively.}
  \label{tab:revisit_memory}
  \wcTableFont
  \setlength{\tabcolsep}{2.0pt}
  \renewcommand{\arraystretch}{1.08}
  \begin{tabular}{lrrrr}
    \toprule
    \textbf{Method} & \textbf{MEt3R}$\downarrow$ & \textbf{LPIPS}$\downarrow$ & \textbf{PSNR}$\uparrow$ & \textbf{SSIM}$\uparrow$ \\
    \midrule
    DreamX-World~\cite{team2026dreamxworld}     & 0.548 & 0.627 & 12.898 & 0.243 \\
    Alaya-EVOKE~\cite{yin2026evoke}                  & 0.414 & 0.565 & 12.332 & 0.290 \\
    HY-WorldPlay~\cite{sun2025worldplay}        & 0.394 & 0.515 & 12.983 & 0.252 \\
    Lyra 2.0~\cite{lyra2026}                   & \wcThird{0.334} & \wcThird{0.487} & \wcThird{14.050} & \wcThird{0.390} \\
    Echo-WM~\cite{zhang2026echowm}             & 0.449 & 0.582 & 12.592 & 0.239 \\
    LingBot-World~2~\cite{gao2026lingbotworld2} & 0.492 & 0.633 & 10.449 & 0.219 \\
    Matrix-Game 3.5~\cite{matrixgame35}        & 0.405 & 0.549 & 12.976 & 0.224 \\
    SANA-WM~\cite{zhu2026sanawm}               & 0.397 & 0.553 & 13.142 & 0.246 \\
    \midrule
    \textbf{WorldCrafter} & \wcSecond{0.166} & \wcSecond{0.255} & \wcSecond{18.016} & \wcSecond{0.517} \\
    \textbf{WorldCrafter-fast} & \wcBest{0.129} & \wcBest{0.186} & \wcBest{20.868} & \wcBest{0.616} \\
    \bottomrule
  \end{tabular}

    \end{minipage}\hfill
    \begin{minipage}[t]{0.49\textwidth}
        \captionsetup{type=table}
        \centering
  \caption{\textbf{Camera-control accuracy.} Results are averaged over 725 generated videos using Sim(3)-aligned estimated and target trajectories. Progressively lighter shades of blue mark the best, second-best, and third-best results, respectively.}
  \label{tab:camera_control}
  \wcTableFont
  \setlength{\tabcolsep}{4.0pt}
  \renewcommand{\arraystretch}{1.08}
  \begin{tabular}{lrrr}
    \toprule
    \textbf{Method} & \textbf{RotErr}$\downarrow$ & \textbf{TransErr}$\downarrow$ & \textbf{CamMC}$\downarrow$ \\
    \midrule
    DreamX-World~\cite{team2026dreamxworld}      & 54.116 & 2.759 & 3.138 \\
    Alaya-EVOKE~\cite{yin2026evoke}                   & 26.042 & 2.042 & 2.199 \\
    HY-WorldPlay~\cite{sun2025worldplay}         & 34.051 & 2.146 & 2.359 \\
    Lyra 2.0~\cite{lyra2026}                    & \wcSecond{16.145} & \wcSecond{1.538} & \wcSecond{1.624} \\
    Echo-WM~\cite{zhang2026echowm}              & 21.455 & 2.072 & 2.189 \\
    LingBot-World~2~\cite{gao2026lingbotworld2} & 30.615 & 2.004 & 2.192 \\
    Matrix-Game 3.5~\cite{matrixgame35}         & 19.881 & 1.920 & 2.028 \\
    SANA-WM~\cite{zhu2026sanawm}                & 23.531 & 1.740 & 1.887 \\
    \midrule
    \textbf{WorldCrafter} & \wcBest{13.536} & \wcBest{1.475} & \wcBest{1.546} \\
    \textbf{WorldCrafter-fast} & \wcThird{18.251} & \wcThird{1.638} & \wcThird{1.737} \\
    \bottomrule
  \end{tabular}

    \end{minipage}
\end{figure*}

\begin{figure*}[!t]
    \centering
    \includegraphics[width=.95\textwidth]{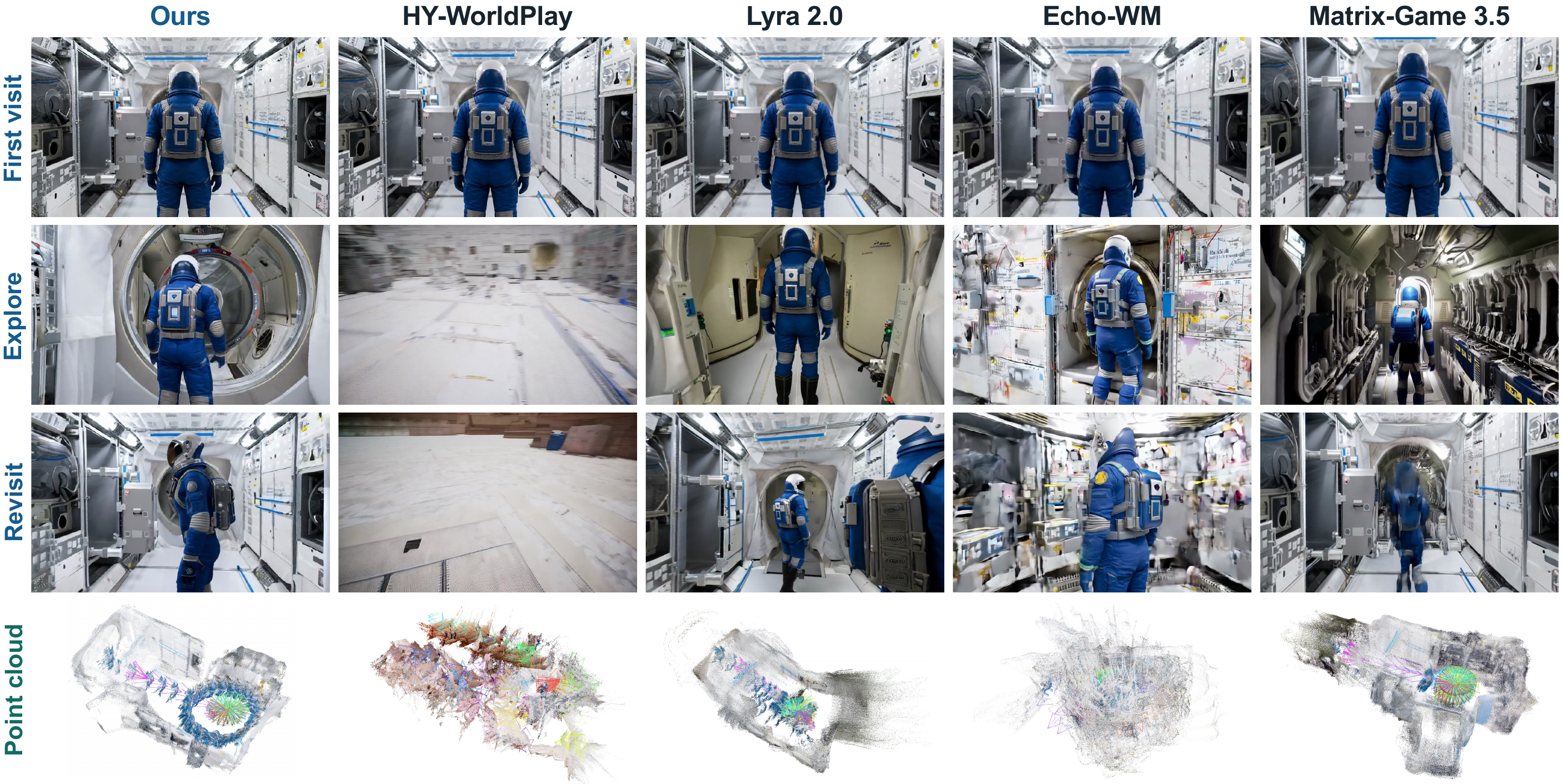}
    \caption{\textbf{Qualitative long-horizon revisit comparison in a dynamic scene.} Rows show the first observation, an intermediate view during exploration, the matched revisit, and the point cloud reconstruction of the generated video, respectively.}
    \label{fig:revisit_qualitative_additional}
    \vspace{0.5em}
    \raggedright
    \begin{minipage}{\textwidth}
        \captionsetup{type=table}
        \centering
  \caption{\textbf{Visual quality on VBench.} Scores are computed in custom-input mode over 725 generated videos. SC, BC, TF, MS, AQ, IQ, DD, and OC denote subject consistency, background consistency, temporal flickering, motion smoothness, aesthetic quality, imaging quality, dynamic degree, and overall consistency, respectively. The Overall column reports the aggregate VBench score. Progressively lighter shades of blue mark the best, second-best, and third-best results, respectively.}
  \label{tab:vbench_results}
  \wcTableFont
  \setlength{\tabcolsep}{3.6pt}
  \renewcommand{\arraystretch}{1.08}
  \begin{tabular}{lrrrrrrrrr}
    \toprule
    \textbf{Method} & \textbf{SC} & \textbf{BC} & \textbf{TF} & \textbf{MS} & \textbf{AQ} & \textbf{IQ} & \textbf{DD} & \textbf{OC} & \textbf{Overall} \\
    \midrule
    DreamX-World~\cite{team2026dreamxworld}      & 75.994 & 87.601 & 95.327 & 98.100 & 57.593 & 60.412 & \wcSecond{99.806} & 24.583 & 77.652 \\
    Alaya-EVOKE~\cite{yin2026evoke}                   & 80.522 & 88.994 & \wcThird{95.872} & 98.481 & \wcThird{61.579} & \wcBest{71.249} & \wcThird{99.029} & 25.015 & \wcSecond{81.406} \\
    HY-WorldPlay~\cite{sun2025worldplay}         & 75.407 & 86.629 & 95.553 & 97.626 & 53.454 & 61.700 & 97.087 & 21.182 & 76.539 \\
    Lyra 2.0~\cite{lyra2026}                    & 77.277 & 87.935 & 95.167 & \wcThird{98.652} & 57.832 & 65.149 & \wcSecond{99.806} & 21.484 & 78.943 \\
    Echo-WM~\cite{zhang2026echowm}              & \wcSecond{81.671} & \wcSecond{90.206} & 95.771 & 96.751 & \wcSecond{61.818} & 66.564 & 98.835 & \wcSecond{26.082} & 80.211 \\
    LingBot-World~2~\cite{gao2026lingbotworld2} & 75.584 & 87.725 & 94.206 & 96.657 & 60.815 & 68.776 & \wcBest{100.000} & \wcThird{25.605} & 78.174 \\
    Matrix-Game 3.5~\cite{matrixgame35}         & 70.370 & 83.626 & 94.395 & 98.087 & 58.615 & \wcSecond{69.374} & \wcBest{100.000} & 18.813 & 76.967 \\
    SANA-WM~\cite{zhu2026sanawm}                & 79.413 & \wcThird{89.625} & 95.768 & 98.608 & \wcBest{63.467} & 66.080 & 98.835 & 25.234 & \wcThird{80.841} \\
    \midrule
    \textbf{WorldCrafter} & \wcBest{82.695} & \wcBest{90.589} & \wcSecond{96.217} & \wcBest{98.787} & 61.432 & \wcThird{69.058} & 96.893 & \wcBest{26.745} & \wcBest{81.910} \\
    \textbf{WorldCrafter-fast} & \wcThird{81.365} & 89.559 & \wcBest{96.444} & \wcSecond{98.736} & 58.650 & 63.982 & 96.505 & 24.306 & 80.285 \\
    \bottomrule
  \end{tabular}

    \end{minipage}
\end{figure*}

\begin{figure*}[!t]
    \begin{minipage}{\textwidth}
        \captionsetup{type=table}
        \centering
    \caption{\textbf{Ablation of the memory design.} Each variant changes one design choice while keeping the training and inference settings and the DiT memory-token budget fixed.}
    \label{tab:ablation_overview}
    \small
    \setlength{\tabcolsep}{4pt}
    \renewcommand{\arraystretch}{1.12}
    \resizebox{.8\textwidth}{!}{%
    \begin{tabular}{lccccccc}
        \toprule
        & \multicolumn{4}{c}{Memory evaluation} & \multicolumn{3}{c}{Camera evaluation} \\
        \cmidrule(lr){2-5}\cmidrule(l){6-8}
        Method & MEt3R$\downarrow$ & LPIPS$\downarrow$ & PSNR$\uparrow$ & SSIM$\uparrow$ & RotErr$\downarrow$ & TransErr$\downarrow$ & CamMC$\downarrow$ \\
        \midrule
        Context memory & 0.382 & 0.497 & 13.907 & 0.315 & 26.522 & 2.083 & 2.150 \\
        Frozen memory encoder & 0.227 & 0.305 & 16.873 & 0.472 & 15.428 & 1.793 & 1.886 \\
        Pose-free memory readout & 0.251 & 0.333 & 16.486 & 0.467 & 18.307 & 1.701 & 1.828 \\
        Similarity-based history retrieval & 0.213 & 0.296 & 17.125 & 0.485 & 14.657 & 1.579 & 1.664 \\
        \midrule
        \rowcolor{wcTableSecond}
        \textbf{WorldCrafter} & \textbf{0.166} & \textbf{0.255} & \textbf{18.016} & \textbf{0.517} & \textbf{13.536} & \textbf{1.475} & \textbf{1.546} \\
        \bottomrule
    \end{tabular}%
    }

    \end{minipage}
    \vspace{0.5em}
    \centering
    \includegraphics[width=.9\textwidth]{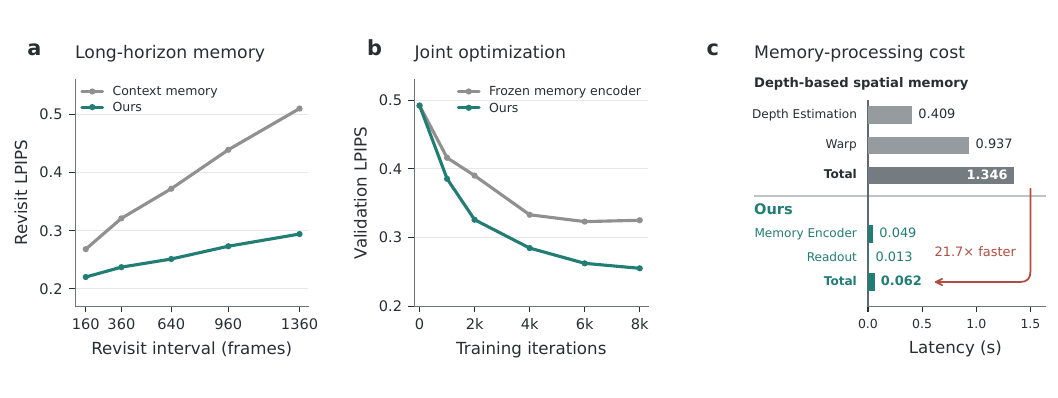}
    \vspace{-2.5em}
    \caption{\textbf{Ablations of implicit 3D-aware memory, joint optimization, and memory-processing efficiency.} (a) Long-horizon revisit consistency of context memory and WorldCrafter's implicit 3D-aware memory, measured by LPIPS across revisit intervals. (b) Validation LPIPS for joint optimization and frozen memory-encoder training at matched iterations. (c) Per-chunk memory-processing latency of depth-based spatial memory and WorldCrafter.}
    \label{fig:ablation_mechanisms}
    \vspace{-1em}
\end{figure*}

{\color{black}
\subsection{Training and Distillation}
\label{sec:dataset_training}
\label{sec:distillation}

\noindent\textbf{Base model training.}
We initialize the video DiT from Helios-base~\cite{yuan2026helios} and train on a subset of OSP dataset~\cite{lin2024open}, DL3DV~\cite{ling2024dl3dv}, and synthetic videos from MIND~\cite{ye2026mind}. Training proceeds through four stages: context-window adaptation, camera-control training, memory-encoder warm-up on VAE latents, and joint optimization. In the final stage, we jointly train the memory encoder, readout module, video DiT, and camera conditioning branch to co-adapt the memory representation and the DiT token space. We first train on real videos and then incorporate synthetic videos to improve dynamic subject modeling.

\noindent\textbf{Distillation.}
We apply distribution matching distillation~\cite{yin2024one,yin2024improved} with a coarse-to-fine pyramid of three spatial resolutions and two denoising steps per resolution. To balance subject-following ability and visual fidelity, we distill high-noise and low-noise models with and without synthetic data, respectively. During inference, we perform 6 denoising steps, using the high-noise model for the first 5 steps and the low-noise model for the final step. This achieves 16 fps on a 4-GPU machine. Full training and distillation details are provided in \hyperref[sec:supp-training]{Supplementary Sec.~\ref*{sec:supp-training}}, and dataset construction is described in \hyperref[sec:supp-dataset]{Supplementary Sec.~\ref*{sec:supp-dataset}}.
}

\section{Experiments}
\label{sec:experiment}

\subsection{Experimental Setup}
\label{sec:experimental_setup}

\noindent\textbf{Benchmark.}
We curate a benchmark to evaluate memory ability, camera-control accuracy, and visual quality in long-horizon video world models. The benchmark contains 145 images from HappyOyster~\cite{happyoyster}, Project Genie~\cite{genie3}, web sources, and images generated by GPT-Image2, covering 83 dynamic object-centric scenes and 62 static scenes. Each image and its text description are paired with 5 metric camera trajectories, yielding 725 videos per method. The trajectories span 528--1,648 frames and include closed-loop revisits to assess whether previously observed content is preserved over long horizons.

\noindent\textbf{Comparison methods.}
We evaluate two variants of our method: WorldCrafter and its distilled counterpart, WorldCrafter-fast. We compare our models with 8 recent camera-controllable video world models: DreamX-World~\cite{team2026dreamxworld}, Alaya-EVOKE~\cite{yin2026evoke}, HY-WorldPlay~\cite{sun2025worldplay}, Lyra 2.0~\cite{lyra2026}, Echo-WM~\cite{zhang2026echowm}, LingBot-World~2~\cite{gao2026lingbotworld2}, Matrix-Game 3.5~\cite{matrixgame35}, and SANA-WM~\cite{zhu2026sanawm}.

These baselines span different memory representations and camera-conditioning mechanisms. HY-WorldPlay and DreamX-World retrieve history context based on camera similarity and use PRoPE~\cite{li2025prope} for camera control. SANA-WM combines Gated DeltaNet memory with UCPE~\cite{zhang2025ucpe} and Pl\"ucker-ray conditioning, whereas Echo-WM employs a UCPE-based camera branch and sliding-window memory. Alaya-EVOKE and Lyra 2.0 construct spatial memory using depth estimated by Depth Anything 3~\cite{lin2025depthanything3} and condition generation through geometric warping. Matrix-Game 3.5 combines geometric patch memory with Warped PRoPE, using VGGT-$\Omega$~\cite{wang2026vggtomega} and Depth Anything 3 for metric-scale geometry annotation.

We evaluate the full-step base model with a refiner for SANA-WM, the full-step base models for Lyra 2.0 and HY-WorldPlay, and the distilled models for the remaining baselines. All methods receive identical initial images, text descriptions, and target trajectories. Before evaluation, we resize the generated videos to $640 \times 384$ to ensure all methods share the same evaluation resolution.

\subsection{Memory Evaluation}
\label{sec:revisit_memory}

Following the protocol in~\cite{hyworld2025}, we evaluate memory ability by comparing frames generated upon revisiting a location with the corresponding frames from the initial visit. We report MEt3R~\cite{asim2025met3r}, LPIPS~\cite{zhang2018lpips}, PSNR, and SSIM~\cite{wang2004ssim} to assess consistency between the paired observations.

As shown in Table~\ref{tab:revisit_memory}, our models achieve the top two results on all 4 metrics, with WorldCrafter-fast performing best. Compared with Lyra 2.0, WorldCrafter reduces LPIPS from 0.487 to 0.255 and increases PSNR from 14.050 to 18.016~dB.
The agreement across these metrics  support the effectiveness of the learned memory representation in long-horizon closed-loop exploration.
Figures~\ref{fig:revisit_qualitative} and~\ref{fig:revisit_qualitative_additional} compare first-visit and revisit frames on static and dynamic scenes, together with point clouds reconstructed from the generated videos using VGGT-$\Omega$~\cite{wang2026vggtomega}. WorldCrafter's revisit frames closely match the corresponding first-visit observations, while its generated videos yield coherent point clouds with well-aligned camera poses. These qualitative results further indicate that WorldCrafter preserves scene structure and appearance during long-horizon video generation while maintaining visual quality.

\subsection{Camera Control Evaluation}
\label{sec:camera_control}

We sample the generated videos at a stride of 4 frames and recover camera trajectories using VGGT-$\Omega$~\cite{wang2026vggtomega}. Following SANA-WM~\cite{zhu2026sanawm}, each trajectory is normalized relative to its first pose and aligned using Umeyama Sim(3) alignment~\cite{umeyama1991least}. Evaluation metrics include rotation error (RotErr), camera-center translation error (TransErr), and pose-matrix discrepancy (CamMC), with lower values indicating more accurate camera control.
As shown in Table~\ref{tab:camera_control}, WorldCrafter achieves the lowest error on all 3 metrics, while WorldCrafter-fast ranks third on each metric.

\subsection{Visual Quality Evaluation}
\label{sec:visual_quality}

Following SANA-WM~\cite{zhu2026sanawm}, we evaluate video generation quality using VBench~\cite{huang2024vbench} in custom-input mode. We report Subject Consistency (SC), Background Consistency (BC), Temporal Flickering (TF), Motion Smoothness (MS), Aesthetic Quality (AQ), Imaging Quality (IQ), Dynamic Degree (DD), and Overall Consistency (OC).
As shown in Table~\ref{tab:vbench_results}, our models achieve the best results in 5 of the 8 dimensions. WorldCrafter obtains the highest overall score of 81.910, while WorldCrafter-fast achieves the best Temporal Flickering score. These gains primarily reflect stronger scene consistency and temporal coherence across the generated videos, complementing the results of the memory evaluation.

\subsection{Ablation Study}
\label{sec:ablation}

All ablations are conducted on WorldCrafter before distillation, following the memory and camera-control evaluation protocols in Secs.~\ref{sec:revisit_memory} and~\ref{sec:camera_control}. Table~\ref{tab:ablation_overview} compares the full model with 4 variants under the same settings.

\noindent\textbf{Implicit 3D-aware memory versus context memory.}
The \textit{context memory} variant replaces the memory tokens produced by the memory encoder and memory readout module with 4 retrieved history latent frames. \textcolor{black}{It is initialized from the camera-conditioned context-memory model before introducing implicit memory (\hyperref[sec:supp-training]{Supplementary Sec.~\ref*{sec:supp-training}}) and further trained for the same number of iterations as the full model.}
Table~\ref{tab:ablation_overview} shows that this replacement degrades both revisit consistency and camera control.
To examine how memory retention varies over time, Figure~\ref{fig:ablation_mechanisms}(a) plots mean revisit LPIPS against the frame interval between the initial observation and its revisit.
WorldCrafter exhibits a slower increase in revisit error, with a widening advantage over context memory at longer intervals.

\noindent\textbf{Joint optimization of the memory encoder and video generator.}
The \textit{frozen memory encoder} variant fixes the encoder during the final training stage, while keeping the memory readout module, video DiT, and camera conditioning branch trainable. This variant yields weaker memory and camera-control performance in Table~\ref{tab:ablation_overview}. Figure~\ref{fig:ablation_mechanisms}(b) tracks validation performance at matched training iterations: joint optimization reaches lower revisit error earlier and maintains this advantage. This supports co-adapting the memory representation with the video generator instead of learning to consume a fixed representation space.

\begin{samepage}
\noindent\textbf{Memory readout.}
The \textit{pose-free memory readout} variant maps the encoded representation to memory tokens without target-pose queries, retaining the same history inputs and output token budget. Table~\ref{tab:ablation_overview} shows that pose-guided readout improves both memory and camera control ability.
\end{samepage}

\noindent\textbf{History retrieval.}
To isolate the effect of history retrieval, the \textit{similarity-based history retrieval} variant replaces max-coverage retrieval with pairwise FoV similarity ranking~\cite{yu2025context,xiao2025worldmem}. Both variants retain the latest latent frame and select 8 additional history frames as the memory encoder input, with the same DiT memory-token budget. Table~\ref{tab:ablation_overview} shows improved revisit consistency and camera control with max-coverage retrieval. This comparison favors selecting complementary views that jointly cover the target region over independently ranking views by similarity, without increasing the number of history inputs.

\noindent\textbf{Memory efficiency.}
Depth-based spatial memory methods, including Lyra 2.0~\cite{lyra2026}, Matrix-Game 3.5~\cite{matrixgame35}, and Alaya-EVOKE~\cite{yin2026evoke}, rely on estimated geometry to reuse history observations. We compare the additional cost of depth estimation and warping with that of WorldCrafter's memory encoder and readout. Figure~\ref{fig:ablation_mechanisms}(c) reports component latencies at $640\times384$ resolution, with a generation chunk of 9 latent frames and 4 history frames for spatial warping. Shared VAE decoding for RGB output and video denoising are excluded. For spatial memory, the depth estimation and alignment process takes 0.409~s using Depth Anything 3~\cite{lin2025depthanything3}, while the batched warping takes 0.937~s, totaling 1.346~s per chunk. By operating directly on history latents, WorldCrafter requires 0.049~s for memory encoding and 0.013~s for readout, totaling 0.062~s. This reduces the summed memory-processing cost by a factor of $21.7\times$, without requiring explicit depth estimation or warping.

\FloatBarrier

\section{Conclusion}
\vspace{-.1em}
\label{sec:discussion}
We present WorldCrafter, a video world model with implicit 3D-aware memory. A learned memory encoder aggregates history latent frames into a compact 3D-aware representation, which is read out as memory tokens that condition on the video DiT to support long-horizon consistency. Through few-step distillation, we further build a real-time system for interactive scene exploration. Experiments demonstrate improved revisit consistency and camera-control accuracy over the evaluated baselines.

\section*{LLM Usage Statement}
The authors used GPT-6 to assist with editing and refining author-produced content and materials related to writing, related work and citations, code and implementation, experiments, figures and visual materials, data preparation, and research ideas, hypotheses, claims, and conclusions. Tool use in each area was limited to an assistive role. The authors did not provide the full manuscript or major portions of it to an LLM provider to obtain reviewer-style feedback. The authors take full responsibility for all submitted content, including its accuracy, originality, and scientific validity.

{
    \small
    \bibliographystyle{ieeenat_fullname}
    \bibliography{main}
}

\clearpage
\maketitlesupplementary
\raggedbottom
\appendix
\setcounter{figure}{0}
\setcounter{table}{0}
\renewcommand{\theHfigure}{supp.\arabic{figure}}
\renewcommand{\theHtable}{supp.\arabic{table}}
\addtocontents{toc}{\protect\setcounter{tocdepth}{2}}
\setcounter{tocdepth}{2}
\tableofcontents
\bigskip

\section{Implementation Details}
\label{sec:supp-implementation}
\subsection{Network Architecture}
\label{sec:supp-network}

\begin{figure*}[t]
\centering
\includegraphics[width=.9\textwidth]{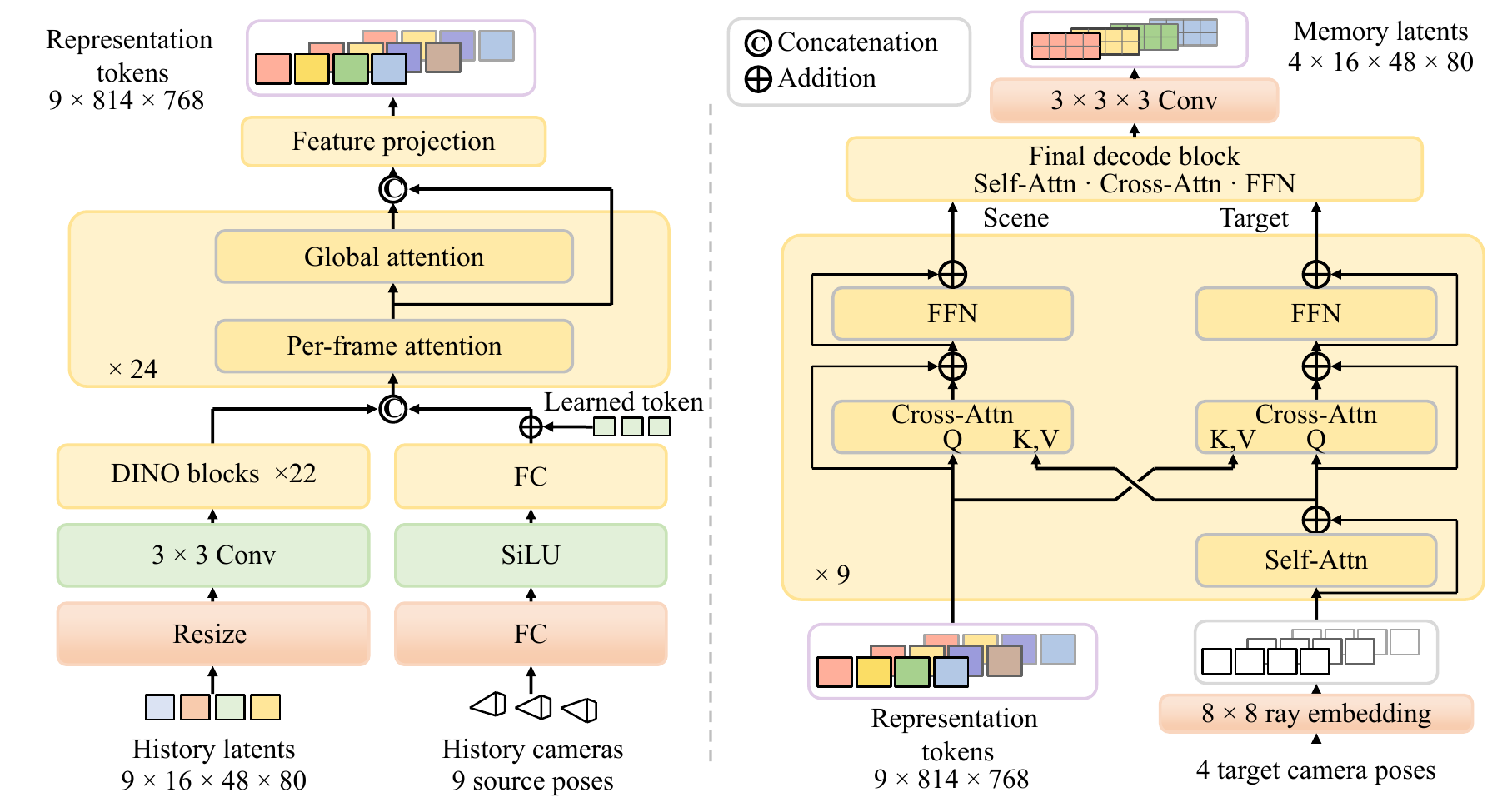}
\caption{\textbf{Memory encoder and pose-guided readout.} The encoder processes selected history latents and their camera tokens to form an implicit 3D-aware representation. Target-camera queries read out 4 memory latent frames for the video DiT. Tensor dimensions omit the batch dimension; latent-frame tensors are shown as frames $\times$ channels $\times$ height $\times$ width.}
\label{fig:supp-memory-encoder}
\end{figure*}

\paragraph{Memory encoder and readout.}
Figure~\ref{fig:supp-memory-encoder} details the memory pathway shared by WorldCrafter and WorldCrafter-fast. The encoder takes the latest history latent frame as a fixed anchor and 8 frames retrieved from the earlier latent history, each with 16 channels and a spatial size of $48\times80$. We resize each latent to $44\times74$ and apply a $3\times3$ convolution with stride 2 to obtain a $22\times37$ grid of 1,024-dimensional patch tokens. These tokens pass through the 22 retained DINO transformer blocks, followed by 24 pairs of per-frame and global attention blocks initialized from LagerNVS~\cite{szymanowicz2026lagernvs}. Each attention block uses 16 heads. Camera tokens encode the source poses, while global attention exchanges information across the selected history frames. The final per-frame and global features are concatenated and projected to 768 dimensions, producing $9\times814$ representation tokens.

The pose-guided readout queries this representation with 4 camera poses sampled at latent-frame indices $\{2,4,6,8\}$ of the upcoming 9-frame chunk, using zero-based indexing. We embed their 6-channel ray maps with an $8\times8$ convolution and process the resulting queries with 10 readout blocks of width 768 and 12 attention heads. A $3\times3\times3$ output convolution maps the queried features to 4 memory latent frames with 16 channels and a spatial size of $48\times80$. This memory is computed once per chunk and reused throughout denoising.

\begin{figure*}[t]
\centering
\includegraphics[width=\textwidth]{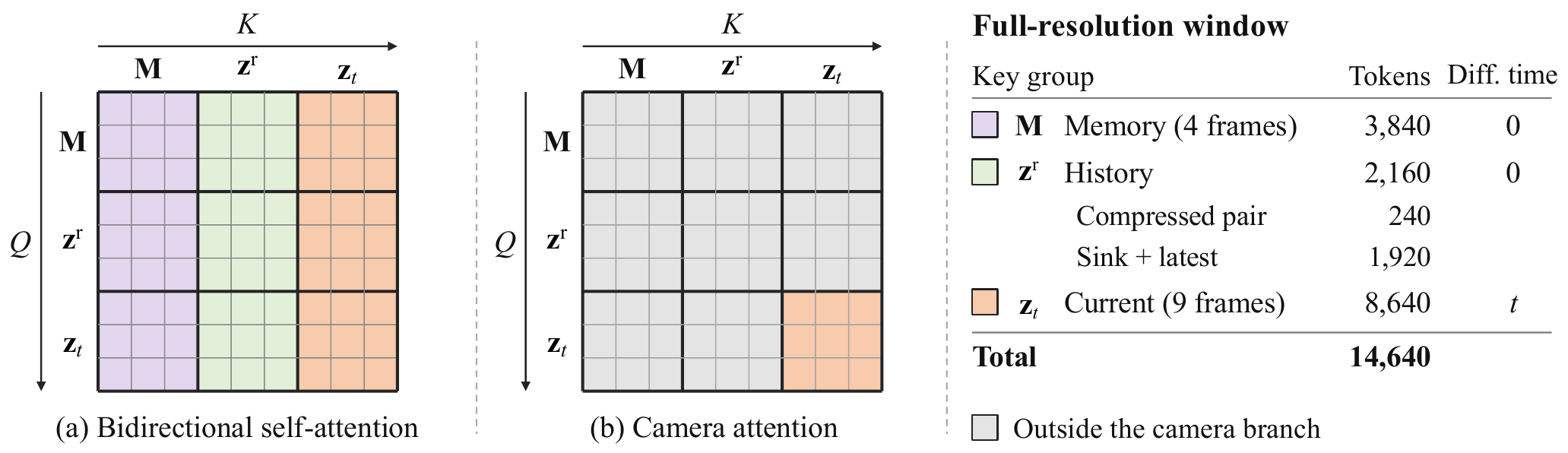}
\caption{\textbf{Attention window for one autoregressive generation step.} (a) Self-attention is bidirectional within the window. (b) The UCPE branch operates only on the current chunk. Rows denote queries and columns denote keys. The right panel lists token counts and diffusion times at full latent resolution.}
\label{fig:supp-attention-window}
\end{figure*}

\paragraph{Attention window.}
WorldCrafter generates chunks autoregressively. Each chunk draws its memory and recent history only from earlier chunks, so generation is causal across chunks. Within the current window, self-attention is bidirectional. As shown in Figure~\ref{fig:supp-attention-window}(a), tokens from the learned memory $\mathbf{M}$, recent history $\mathbf{z}^{\mathrm{r}}$, and noisy current chunk $\mathbf{z}_t$ attend to one another as the current chunk is denoised jointly. Figure~\ref{fig:supp-attention-window}(b) shows the parallel UCPE branch~\cite{zhang2025ucpe}, which computes camera attention only among current-chunk tokens. Memory and history tokens are assigned diffusion time 0, and current tokens are assigned time $t$. Once denoised, the current chunk becomes part of the history for the next chunk.
 As shown in the right panel of Figure~\ref{fig:supp-attention-window}, at a latent resolution of $48\times80$ the 4 memory frames, retained history, and 9 current frames contribute 3,840, 2,160, and 8,640 tokens, respectively, for a total of 14,640.

\paragraph{History retrieval algorithm.}
For the encoder input of $k=9$ history latent frames, we fix the latest frame as the anchor. Algorithm~\ref{alg:history-retrieval} retrieves the remaining $k-1=8$ frames from the earlier history to cover the 4 query poses $\mathbf C^{\mathrm q}$ along the upcoming camera trajectory. At each step, we evaluate the coverage of each target view after adding a candidate to the anchor and previously retrieved frames. We compare candidates by their lowest resulting coverage, then by their second-lowest coverage, and so on. If all four values tie, we give priority to regions newly entering the field of view at the final target pose and use recency only if the geometric criteria still tie.

Following the frustum sampling scheme of Matrix-Game 3.0~\cite{matrixgame3}, we sample a $10\times10\times10$ grid at depths from 0.1 to 30\,m in each target camera frustum. We use horizontal and vertical fields of view of $100^\circ$ and $71.13^\circ$, respectively. A target view's coverage is the fraction of its sampled points that fall within the field of view of at least one selected history camera. As shown in \hyperref[tab:ablation_overview]{Table~\ref*{tab:ablation_overview}} of the main paper (\hyperref[sec:ablation]{Sec.~\ref*{sec:ablation}}), max-coverage retrieval improves revisit consistency and camera control over pairwise FoV similarity ranking with the same number of history frames.

\begin{algorithm}[t]
\caption{Max-coverage history retrieval}
\label{alg:history-retrieval}
\small
\algrenewcommand\algorithmicrequire{\textbf{Input:}}
\algrenewcommand\algorithmicensure{\textbf{Output:}}
\begin{algorithmic}
\Require History latents $\mathbf z^{\mathrm h}$ and camera parameters $\mathbf C^{\mathrm h}$ in chronological order, and 4 query poses $\mathbf C^{\mathrm q}$.
\Ensure The anchor and 8 retrieved frames, with latents $\mathbf z^{\mathrm s}$ and camera parameters $\mathbf C^{\mathrm s}$.
\end{algorithmic}
\smallskip
\noindent\textbf{Notation}\par\nobreak
{\renewcommand{\arraystretch}{1.06}
\noindent\begin{tabular}{@{}l@{\hspace{0.7em}}p{.77\linewidth}@{}}
$i,\,j$ & History frame index and query view index. History ends at frame $n$ (the anchor), and $j=1,\ldots,4$.\\
$P_j$ & Grid of 3D samples in the frustum of query view $j$.\\
$V_{ij}$ & Samples in $P_j$ inside history camera $i$'s field of view.\\
$U_j$ & Samples in $P_j$ covered by frames selected so far.\\
$S,\,E$ & Selected frame indices (including the anchor) and eligible candidate indices for the current step.\\
$c_{ij}$ & Fraction of $P_j$ covered after adding candidate $i$.\\
$s_i,\,i^*$ & Candidate $i$'s four coverage fractions sorted from lowest to highest, and the winning candidate index.\\
\end{tabular}\par}
\smallskip
\begin{algorithmic}[1]
\State Sample $P_j$ and compute $V_{ij}$ for all $i,j$.
\State $S\gets\{n\}$ and $U_j\gets V_{nj}$ for each $j$.
\For{8 retrieval steps}
    \State $E\gets\{i\notin S : V_{ij}\cap U_j\ne\varnothing\text{ for some }j\}$
    \If{$E=\varnothing$} \State $E\gets\{1,\ldots,n\}\setminus S$ \EndIf
    \For{$i\in E$}
        \State $c_{ij}\gets |U_j\cup V_{ij}|/|P_j|$ for each $j$.
        \State $s_i\gets\operatorname{sort}_{\uparrow}(c_{i1},c_{i2},c_{i3},c_{i4})$
    \EndFor
    \State Compare the scores $s_i$ from left to right, preferring the larger value at the first difference.
    \State $S\gets S\cup\{i^*\}$ and $U_j\gets U_j\cup V_{i^*j}$ for each $j$.
\EndFor
\State \Return $(\mathbf z^{\mathrm s},\mathbf C^{\mathrm s})$ at the selected indices $S$.
\end{algorithmic}
\end{algorithm}

\subsection{Hyper-parameters}
\label{sec:supp-hyperparameters}
As shown in Table~\ref{tab:supp-hyperparameters}, WorldCrafter-fast retains the same memory pathway and context-window structure as WorldCrafter. Its principal inference differences are distilled sampling, a coarse-to-fine spatial pyramid, and a hybrid high-noise/low-noise denoiser.

\subsection{Training and Distillation Details}
\label{sec:supp-training}

\noindent\textbf{Base model training.}
We initialize the video DiT from Helios-base~\cite{yuan2026helios} and train our base model in 4 stages. The original Helios-base inference window contains a 9-frame noise chunk and a FramePack-style clean history~\cite{zhang2025framepack} comprising a compressed 16-frame segment, a 2-frame segment, the latest latent frame, and an attention-sink frame. We remove its compressed 16-frame segment and prepend memory tokens equivalent in number to the tokens of 4 uncompressed history frames.

In the first stage, we fine-tune Helios-base to adapt to our modified inference window. For each sampled video chunk, we randomly select 4 history frames from the preceding 4 chunks to populate the memory slots. We train on 760,000 videos from the OSP dataset for 5,000 iterations using 32 GPUs with a global batch size of 32.

In the second stage, we introduce camera control by training the UCPE-based camera conditioning branch while keeping the video DiT backbone frozen. We use 40,000 videos from the filtered OSP subset and 6,000 videos from DL3DV, training on 32 GPUs with a global batch size of 128.

\begin{table*}[t]
\centering
\caption{\textbf{Architecture and inference hyperparameters.} Sampling settings refer to image-conditioned exploration at $384\times640$ output resolution. With no history, WorldCrafter-fast uses 12 steps (4 per pyramid level).}
\label{tab:supp-hyperparameters}
\small
\setlength{\tabcolsep}{8pt}
\renewcommand{\arraystretch}{1.07}
\begin{tabular}{@{}lll@{}}
\toprule
\textbf{Architecture parameter} & \multicolumn{2}{l@{}}{\textbf{Shared by WorldCrafter and WorldCrafter-fast}} \\
\midrule
\multicolumn{3}{@{}l}{\textit{Video generator and attention window}}\\
Video DiT & \multicolumn{2}{l@{}}{40 blocks, width 5,120, 40 heads} \\
VAE latent channels & \multicolumn{2}{l@{}}{16} \\
Full latent resolution & \multicolumn{2}{l@{}}{$48\times80$} \\
Current chunk length & \multicolumn{2}{l@{}}{9 latent frames} \\
Memory size & \multicolumn{2}{l@{}}{4 latent frames; 3,840 full-resolution tokens} \\
History window & \multicolumn{2}{l@{}}{2 compressed frames + latest frame + sink frame} \\
Memory / current / short-history patch & \multicolumn{2}{l@{}}{$1\times2\times2$} \\
Compressed-history patch & \multicolumn{2}{l@{}}{$2\times4\times4$} \\
Camera conditioning & \multicolumn{2}{l@{}}{Parallel UCPE branch on current tokens} \\
\midrule
\multicolumn{3}{@{}l}{\textit{Memory encoder and readout}}\\
Encoder input & \multicolumn{2}{l@{}}{8 retrieved frames + latest history frame} \\
Patch embedding & \multicolumn{2}{l@{}}{$3\times3$ convolution, stride 2; $22\times37$ tokens/frame} \\
DINO transformer & \multicolumn{2}{l@{}}{22 blocks, width 1,024, 16 heads} \\
Per-frame and global attention & \multicolumn{2}{l@{}}{24 block pairs, width 1,024, 16 heads} \\
Representation tokens & \multicolumn{2}{l@{}}{$9\times814$ tokens, width 768} \\
Pose-guided readout & \multicolumn{2}{l@{}}{10 blocks, width 768, 12 heads} \\
Query indices within current chunk & \multicolumn{2}{l@{}}{$\{2,4,6,8\}$, zero-based} \\
Ray embedding & \multicolumn{2}{l@{}}{$8\times8$ convolution, stride 8; 6 input channels} \\
Output projection & \multicolumn{2}{l@{}}{$3\times3\times3$ convolution, $768\rightarrow16$ channels} \\
\midrule
\textbf{Inference parameter} & \textbf{WorldCrafter} & \textbf{WorldCrafter-fast} \\\midrule
Output resolution  & $384\times640$  & $384\times640$  \\
Denoising steps with history & 50 & 6 ($2+2+2$) \\
Number of spatial denoising levels & 1 & 3 \\
Latent denoising resolutions & $48\times80$ & $12\times20\rightarrow24\times40\rightarrow48\times80$ \\
Classifier-free guidance scale & 5.0 & 1.0 \\
Denoiser choice (I2V) & Base model & High-noise (steps 1--5), low-noise (step 6) \\
\bottomrule
\end{tabular}
\end{table*}

In the third stage, we adapt our memory encoder to process VAE latents. We initialize it from the LagerNVS encoder and replace its shallow DINO layers with a latent patch embedding layer, as described in Sec.~\ref{sec:supp-network}. The encoder takes a fixed number of 9 latent frames as input. We warm up the memory encoder on DL3DV and the filtered OSP subset for 5,000 iterations using 16 GPUs with a global batch size of 16.

In the final stage, we introduce a memory readout module that outputs a fixed number of memory tokens matching the token count of 4 full frames. We jointly train this module with the memory encoder, video DiT, and camera conditioning branch to co-adapt the learned memory representation and the DiT token space. We first train on DL3DV and the filtered OSP subset for 8,000 iterations using 32 GPUs with a global batch size of 32, then incorporate synthetic videos from MIND for 1,000 additional iterations to improve dynamic subject modeling.

\noindent\textbf{Pyramid distillation.}
Following Helios~\cite{yuan2026helios}, we adopt a coarse-to-fine pyramid denoising scheme and apply distribution matching distillation~\cite{yin2024one,yin2024improved} to reduce the number of sampling steps. We use 3 spatial resolutions with 2 denoising steps per resolution. To support camera conditioning across the pyramid, we rescale the spatial coordinates while keeping the camera poses and field of view unchanged across pyramid levels during UCPE camera embedding.

\noindent\textbf{Hybrid distilled model.}
We observe a trade-off between visual fidelity and subject-following ability when distilling with synthetic data. Incorporating synthetic data~\cite{ye2026mind} improves the model's subject-following ability, but can also introduce smeared textures. To preserve both subject-following ability and natural visual details, we distill a low-noise model and a high-noise model with different training data compositions. The low-noise model is distilled from the base model before synthetic data adaptation, using the filtered OSP subset and DL3DV dataset to preserve natural appearance. The high-noise model is distilled from the base model after synthetic data adaptation, using a mixture of OSP, DL3DV, and MIND to retain subject-following ability. During inference, the low-noise model performs the last denoising step, while the high-noise model performs all the preceding steps.
After distillation, WorldCrafter-fast can achieve a generation speed of 16 fps on a 4-GPU machine.

\section{Additional Evaluation}
\label{sec:supp-evaluation}
\begin{table*}[t]
\centering
\caption{\textbf{Evaluation on the ORBIA benchmark.}  All scores are on a 0--100 scale ($\uparrow$). Input Pres., Gen. Persist., 3D Consist., and Temp. Stab. denote input preservation, generated-content persistence, 3D self-consistency, and temporal stability. Interface indicates the supplied control signal. The \colorbox{wcTableBest}{\strut best}, \colorbox{wcTableSecond}{\strut second-best}, and \colorbox{wcTableThird}{\strut third-best} results are highlighted per column.}
\label{tab:supp-orbia}
\small
\setlength{\tabcolsep}{4.4pt}
\renewcommand{\arraystretch}{1.08}
\begin{tabular*}{\textwidth}{@{\extracolsep{\fill}}llrrrrrrr@{}}
\toprule
Model & Interface & Overall & Control & Quality & Input Pres. & Gen. Persist. & 3D Consist. & Temp. Stab. \\
\midrule
\textbf{WorldCrafter-fast (ours)} & Camera & \wcBest{80.50} & \wcBest{85.79} & 73.26 & \wcBest{78.21} & \wcBest{84.77} & 70.72 & \wcBest{86.85} \\
\midrule
SolarWM-H3~\cite{huang2026solarwmopendatascalable} & Camera & \wcSecond{76.32} & \wcThird{79.93} & 74.76 & 69.26 & \wcSecond{70.28} & 68.84 & \wcSecond{84.05} \\
Lyra 2.0~\cite{lyra2026} & Camera & \wcThird{74.77} & \wcSecond{81.64} & 71.37 & 68.79 & \wcThird{70.25} & 61.51 & 82.31 \\
Alaya-EVOKE~\cite{yin2026evoke} & Camera & 72.70 & 66.47 & 76.48 & 68.81 & 66.43 & 68.02 & \wcThird{83.18} \\
Echo-WM~\cite{zhang2026echowm} & Camera & 72.24 & 70.74 & 75.02 & 65.91 & 61.42 & 65.26 & 82.69 \\
Gen3C~\cite{ren2025gen3c} & Camera & 71.95 & 75.51 & 67.55 & \wcSecond{75.23} & 68.58 & 60.16 & 78.96 \\
Matrix-Game 3.5~\cite{matrixgame35} & Camera & 68.07 & 64.18 & 74.03 & 66.93 & 60.82 & 54.61 & 76.41 \\
Seedance2.5~\cite{bytedance2026seedance25} & Text & 67.44 & 43.23 & 76.97 & \wcThird{72.84} & 67.70 & \wcSecond{72.54} & 80.39 \\
AlayaWorld~\cite{alayaworld2026} & Camera & 67.01 & 61.68 & 71.04 & 70.17 & 56.62 & 58.92 & 76.31 \\
Genie3~\cite{genie3} & Action & 66.72 & 55.26 & 74.43 & 61.01 & 54.41 & 61.25 & 83.15 \\
LingBot-World 2~\cite{gao2026lingbotworld2} & Camera & 65.98 & 58.48 & 75.34 & 61.54 & 52.00 & 57.11 & 77.28 \\
HY-WorldPlay~\cite{sun2025worldplay} & Camera & 65.50 & 45.26 & 72.06 & 66.84 & 69.84 & 70.34 & 77.32 \\
Kling3.0~\cite{kuaishou2026kling3} & Text & 65.23 & 38.58 & 77.38 & 69.14 & 63.32 & 71.30 & 79.30 \\
SANA-WM~\cite{zhu2026sanawm} & Camera & 65.08 & 59.39 & 73.22 & 63.19 & 51.95 & 54.10 & 75.03 \\
HappyOyster~\cite{happyoyster} & Action & 64.34 & 60.49 & 68.78 & 58.23 & 46.31 & 59.17 & 78.25 \\
Cosmos3-Super~\cite{nvidia2026cosmos3reference} & Text & 63.72 & 27.92 & \wcBest{79.04} & 66.04 & 69.75 & \wcBest{74.55} & 79.73 \\
HunyuanVideo1.5~\cite{team2025hunyuanvideo15} & Text & 63.27 & 36.09 & 77.07 & 63.12 & 57.44 & 71.31 & 78.96 \\
Cosmos3-Nano~\cite{nvidia2026cosmos3reference} & Camera & 62.75 & 35.70 & 74.25 & 66.78 & 66.29 & 70.10 & 74.71 \\
LingBot-World 1~\cite{team2026lingbotworld} & Camera & 62.73 & 49.12 & 73.84 & 59.56 & 50.71 & 58.93 & 75.37 \\
Voyager~\cite{huang2025voyager} & Camera & 62.72 & 60.92 & 68.05 & 58.65 & 43.72 & 46.32 & 78.04 \\
MiniMax-H3~\cite{minimax2026h3} & Text & 62.00 & 32.48 & \wcSecond{78.09} & 63.36 & 55.79 & 69.42 & 77.51 \\
DreamX-World~\cite{team2026dreamxworld} & Camera & 61.79 & 48.99 & 74.96 & 60.96 & 48.70 & 48.73 & 74.82 \\
LTX2.5~\cite{lightricks2026ltx25} & Text & 61.25 & 26.51 & \wcThird{77.96} & 64.81 & 60.25 & \wcThird{71.64} & 77.30 \\
FantasyWorld~\cite{dai2025fantasyworld} & Camera & 60.64 & 39.16 & 70.70 & 63.08 & 60.17 & 63.53 & 72.51 \\
Wan2.7~\cite{alibaba2026wan27} & Text & 60.55 & 31.31 & 76.46 & 65.96 & 54.35 & 64.19 & 75.79 \\
ABot-World~\cite{jiang2026abotworld} & Action & 58.87 & 47.06 & 66.97 & 57.47 & 45.22 & 48.32 & 76.32 \\
LongCat~\cite{team2025longcatvideo} & Text & 57.18 & 23.76 & 77.30 & 58.89 & 47.77 & 64.11 & 74.22 \\
Astra~\cite{zhu2025astra} & Camera & 52.30 & 28.45 & 61.30 & 59.38 & 44.74 & 59.91 & 67.31 \\
\bottomrule
\end{tabular*}
\end{table*}

\subsection{Dataset and Metrics}
\label{sec:supp-orbia-protocol}

We further assess WorldCrafter-fast on the ORBIA benchmark~\cite{lu2026orbia}, which evaluates whether generated views preserve a consistent scene during camera-controlled exploration. ORBIA contains 720 cases drawn from ScanNet++~\cite{yeshwanth2023scannetpp}, SpatialVID~\cite{wang2026spatialvid}, Sekai~\cite{li2025sekai}, and MiraData~\cite{ju2024miradata}, as well as videos recorded by the ORBIA authors and scenes they rendered in Unreal Engine. These comprise 480 short trajectories of approximately 22 seconds and 240 minute-long extensions, with one and three revisits, respectively. Six trajectory templates cover forward--backward motion, lateral return, yaw return, arc return, orbit around an anchor, and peek-and-return. Each case provides an initial image, a scene caption, and a target trajectory, together with registered RGB-D observations and annotated landmarks for evaluation.

The benchmark distinguishes three aspects of consistency. \emph{Input preservation} measures whether the generated query and return views retain the appearance and geometry of input-visible surfaces and landmarks. \emph{Generated-content persistence} evaluates whether newly synthesized regions remain consistent upon revisiting, using their first generated observations as references. \emph{3D self-consistency} measures agreement among generated views through depth and feature correspondences, regional scale agreement, and held-out rendering from a reconstructed Gaussian scene. Depth Anything 3 recovers camera poses and depth for these geometric comparisons. ORBIA additionally measures \emph{control alignment}, \emph{visual quality}, and \emph{temporal stability}; the latter tracks quality, control, and local geometry in five-second windows. Scores use a 0--100 scale ($\uparrow$). Overall weights are 25\% each for control and quality, 10\% for each consistency dimension, and 20\% for temporal stability.

\subsection{Baselines}
\label{sec:supp-orbia-baselines}

\noindent\textbf{Camera-conditioned models.}
The 16 baselines that accept camera parameters are SolarWM-H3~\cite{huang2026solarwmopendatascalable}, Lyra 2.0~\cite{lyra2026}, Alaya-EVOKE~\cite{yin2026evoke}, Echo-WM~\cite{zhang2026echowm}, Gen3C~\cite{ren2025gen3c}, Matrix-Game 3.5~\cite{matrixgame35}, AlayaWorld~\cite{alayaworld2026}, LingBot-World 1~\cite{team2026lingbotworld}, LingBot-World 2~\cite{gao2026lingbotworld2}, HY-WorldPlay~\cite{sun2025worldplay}, SANA-WM~\cite{zhu2026sanawm}, Cosmos3-Nano~\cite{nvidia2026cosmos3reference}, Voyager~\cite{huang2025voyager}, DreamX-World~\cite{team2026dreamxworld}, FantasyWorld~\cite{dai2025fantasyworld}, and Astra~\cite{zhu2025astra}. ORBIA uses the distilled checkpoints for Lyra 2.0 and HY-WorldPlay and the streaming checkpoint with a refiner for SANA-WM.

\noindent\textbf{Action-conditioned models.}
For Genie3~\cite{genie3}, HappyOyster~\cite{happyoyster}, and ABot-World~\cite{jiang2026abotworld}, ORBIA converts the target trajectory into discrete movement and look commands.

\noindent\textbf{Text-conditioned models.}
For Seedance2.5~\cite{bytedance2026seedance25}, Kling3.0~\cite{kuaishou2026kling3}, Cosmos3-Super~\cite{nvidia2026cosmos3reference}, HunyuanVideo1.5~\cite{team2025hunyuanvideo15}, MiniMax-H3~\cite{minimax2026h3}, LTX2.5~\cite{lightricks2026ltx25}, Wan2.7~\cite{alibaba2026wan27}, and LongCat~\cite{team2025longcatvideo}, ORBIA describes the requested camera motion in the text prompt.

\subsection{Results}
\label{sec:supp-orbia-results}

As shown in Table~\ref{tab:supp-orbia}, WorldCrafter-fast achieves the highest overall score among the compared methods on ORBIA, scoring 80.50 compared with 76.32 for SolarWM-H3. It also achieves the highest scores in control alignment, input preservation, generated-content persistence, and temporal stability. The largest margin is in generated-content persistence: WorldCrafter-fast scores 84.77, compared with 70.28 for the strongest baseline. This result directly reflects the ability to retain newly generated content after the camera moves away and later revisits it. Together with its input-preservation score of 78.21 and temporal-stability score of 86.85, these results support the effectiveness of implicit 3D-aware memory across exploration and repeated revisits.

\section{Additional Ablations}
\par\penalty0
\label{sec:supp-ablations}

\begin{figure*}[t]
\centering
\includegraphics[width=.9\textwidth]{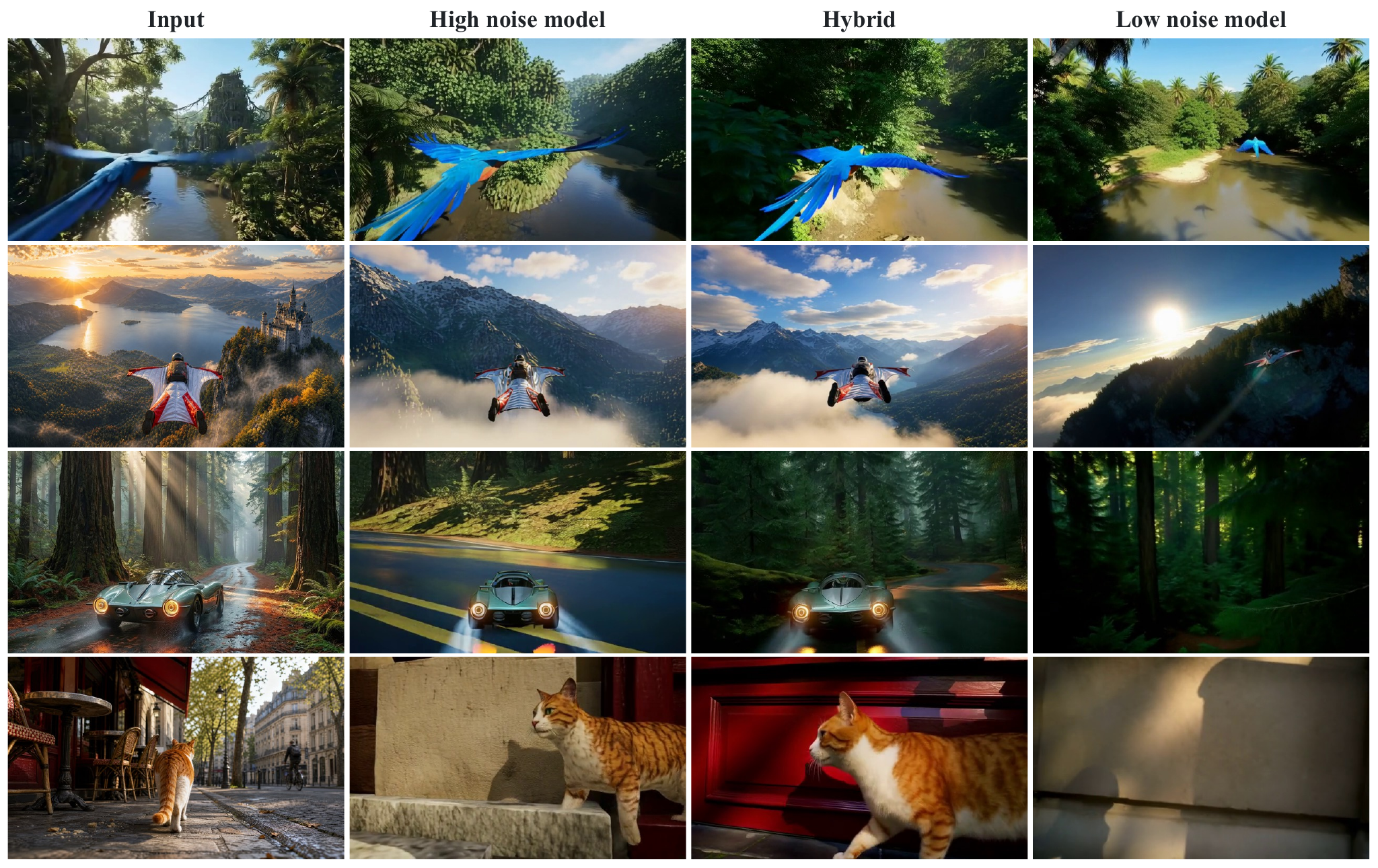}
\caption{\textbf{Effect of the denoising schedule on subject following and visual detail.} Each row shows an input image followed by generated frames from the high-noise model alone, the hybrid schedule, and the low-noise model alone, from left to right.}
\label{fig:supp-hybrid-distillation}
\end{figure*}

\noindent\textbf{Effectiveness of the hybrid distilled model.}
Figure~\ref{fig:supp-hybrid-distillation} compares the two distilled models with our hybrid schedule. The high-noise model, trained on a mixture of real and synthetic videos, follows moving subjects more reliably as the viewpoint changes. Its textures, however, often lack fidelity and can drift over time. The low-noise model is trained exclusively on real videos and produces more natural textures with less visual drift. Its weaker subject-following ability is particularly apparent in third-person scenes, where a change in viewpoint can cause the camera to lose track of the subject. WorldCrafter-fast therefore uses the high-noise model for the first five denoising steps to establish subject motion and framing, then uses the low-noise model in the final step to refine texture and appearance, as described in Sec.~\ref{sec:supp-training}.

\section{Details of the Constructed Dataset}
\par\penalty0
\label{sec:supp-dataset}

Our training data combine the Open-Sora-Plan (OSP) dataset~\cite{lin2024open}, DL3DV~\cite{ling2024dl3dv}, and synthetic videos from MIND~\cite{ye2026mind}, covering diverse indoor and outdoor scenes and object motions. We use Depth Anything 3~\cite{lin2025depthanything3} to obtain metric-scale camera pose annotations across all data sources and Qwen2.5-VL~\cite{bai2025qwen25vl} to generate video captions.

\noindent\textbf{Subject-following videos.}
We select OSP videos whose captions describe the camera tracking a moving subject. Within these videos, we screen each training window for sufficient camera motion. For each window, we measure the displacement of the camera center and the 3D rotation angle relative to the first frame. We denote their respective maxima over valid camera poses by $t_{\mathrm{peak}}$ and $r_{\mathrm{peak}}$. A window passes the motion check if its peak displacement satisfies $t_{\mathrm{peak}}\leq10\,\mathrm{m}$ and either $r_{\mathrm{peak}}\geq15^\circ$ or both $t_{\mathrm{peak}}\geq0.8\,\mathrm{m}$ and $r_{\mathrm{peak}}\geq5^\circ$. This admits both turns in place and tracking shots with more modest rotation.

\noindent\textbf{Trajectory smoothness.}
A single erroneous pose estimate can produce a large peak displacement or rotation. We therefore check the smoothness of every trajectory that passes the motion threshold. The cumulative path length must be at most twice the straight-line distance between the first and last camera centers. For successive translation directions, the maximum and mean turning angles must be at most $120^\circ$ and $60^\circ$, respectively. The largest translation step must be at most four times the median step length, and the rotation between consecutive valid poses must be at most $75^\circ$. These criteria favor smooth camera motion and can also exclude genuine sharp turns or reversals. We further narrow the training pool by subject matter. As detailed in Sec.~\ref{sec:supp-training}, we use 40,000 videos from the filtered OSP subset during camera-control.

\section{Interactive Demo}
\label{sec:supp-demo}
\begin{figure*}[!t]
\centering
\includegraphics[width=\linewidth]{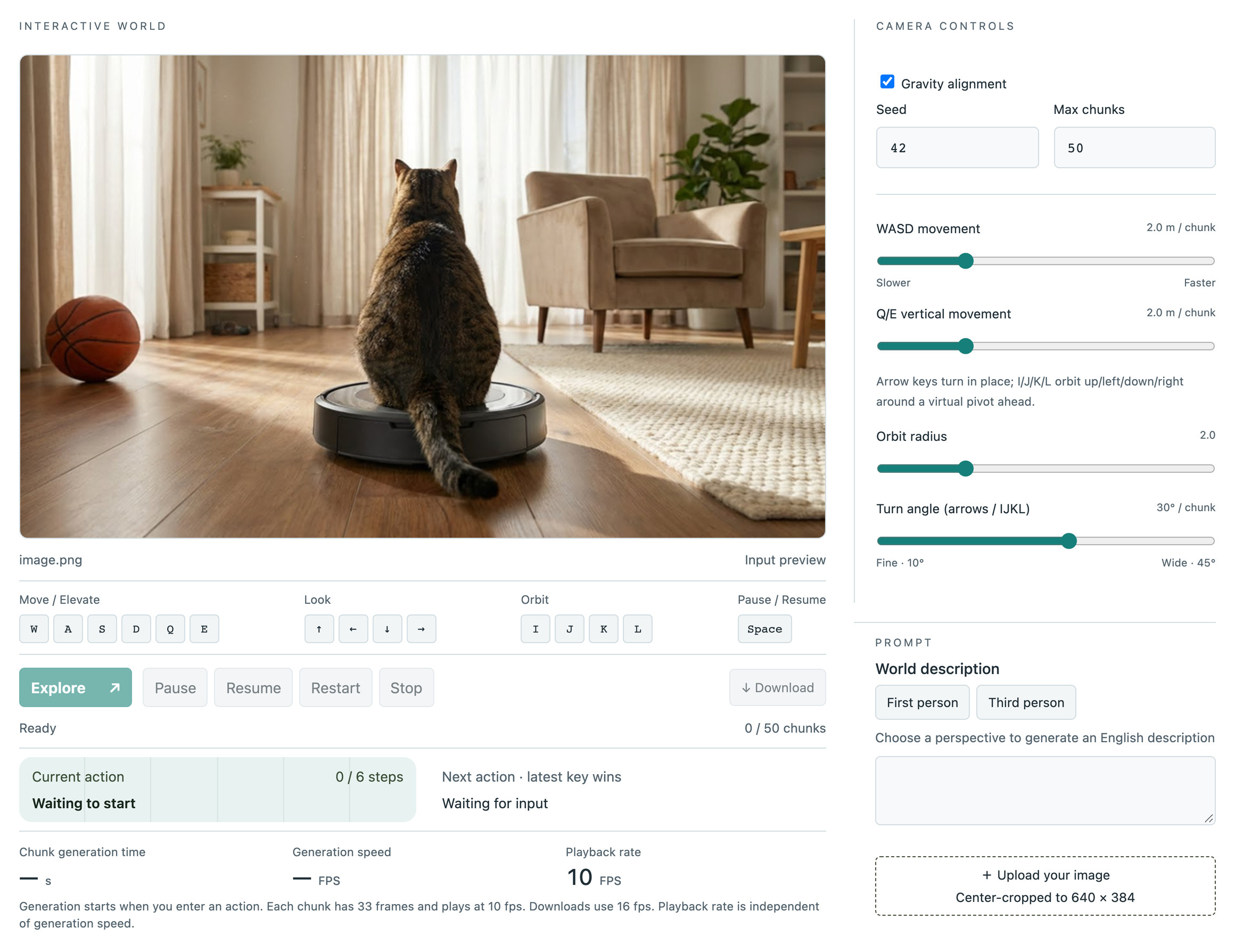}
\caption{\textbf{Interactive exploration interface.} The interface is shown in the uploaded-image preview state, before captioning and generation. The left panel contains the viewport, keyboard controls, session actions, and generation status. The right panel provides gravity alignment, camera-motion parameters, and automatic captioning with first- and third-person options. Generation timing is reported separately from playback rate.}
\label{fig:supp-demo-interface}
\end{figure*}

\subsection{System Overview}
\label{sec:supp-demo-system}
We deploy WorldCrafter-fast as a browser-based exploration system in which users choose camera actions while viewing the generated scene. Each action advances a persistent session. The browser handles input and playback, while a server coordinates camera trajectories and inference. A dedicated worker keeps the model resident and retains latent history and camera state across chunks. Controls and progress notifications are exchanged over WebSocket, and completed video segments are served over HTTP. This separation allows the interface to receive new input while generation is in progress. Figure~\ref{fig:supp-demo-interface} shows the browser interface, including the scene viewport and camera controls.

\subsection{User Interface and Camera Controls}
\label{sec:supp-demo-controls}
\noindent\textbf{Scene initialization and automatic captioning.}
Users can select a preset with an accompanying description or upload an image, which is center-cropped and resized to $640\times384$. For uploaded images, Qwen3-VL-4B-Instruct~\cite{bai2025qwen3vl} generates an English description using a first- or third-person prompt template. The first-person template emphasizes the visible scene layout and appearance; the third-person template describes a visible subject and a continuous action appropriate to that subject, falling back to a scene description when no suitable subject is present. Users can edit the resulting caption before starting exploration. These options determine the text conditioning, while camera motion remains controlled through the same keyboard interface. The panel also exposes the random seed and maximum number of chunks.

\noindent\textbf{Camera reference frames and motion controls.}
Camera poses follow a camera-to-world convention with local axes pointing right, down, and forward. The system retains a common trajectory relative to the initial camera for history retrieval and computes chunk-relative poses for camera conditioning. WASD moves along the current camera's forward or right direction projected onto the horizontal plane, so looking up or down does not change the height of horizontal movement. Q/E moves along the world vertical axis. The arrow keys rotate the camera in place: yaw is defined about the world vertical axis, whereas pitch uses the camera-local right axis. I/J/K/L moves the camera around a virtual pivot ahead of it while maintaining the orbit radius. Consecutive orbit actions preserve this pivot; translation, in-place rotation, or a radius change establishes a new pivot from the updated pose. Sliders independently control horizontal and vertical translation over $1$--$5$\,m per chunk, rotation over $10^{\circ}$--$45^{\circ}$ per chunk, and orbit radius over $1$--$5$\,m.

\noindent\textbf{Gravity alignment.}
An input image may have nonzero camera pitch or roll, making its image axes unsuitable as a level navigation frame. With gravity alignment enabled, GeoCalib~\cite{veicht2024geocalib} estimates the gravity direction from the input image when the user starts exploration. The estimated direction defines the vertical axis for translation, yaw, and horizontal orbit, while pitch and vertical orbit retain their camera-local rotation axis. This changes the control reference frame without rotating the input image or changing the generation field of view.

\noindent\textbf{Session controls and feedback.}
The interface displays the active and queued actions. Users can pause after the current chunk, resume from the saved history, or restart with the same input and settings. Esc or loss of window focus clears queued movement. Users can download the video, and the server saves the camera trajectory and session metadata.

\subsection{Streaming Generation}
\label{sec:supp-demo-streaming}
A key press requests one 33-frame chunk at $640\times384$. Holding the key repeats the action. Once a chunk starts, its camera trajectory is fixed and new input controls the next chunk. For each subsequent chunk, the model reads pose-conditioned memory from the retrieved history before denoising, then adds the generated latents to history.

The server streams each completed chunk as an MP4 segment. The browser preloads the next segment during playback and holds the last frame if generation is still in progress. Playback runs at 10\,fps and downloads retain the 16\,fps encoding rate.

\section{Limitations}
\par\penalty0
\label{sec:supp-limitations}

Although WorldCrafter improves revisit consistency and supports real-time exploration, several limitations remain. First, the distilled model can still exhibit visual drift during prolonged exploration. Because each generated chunk becomes part of the history used to generate later chunks, small errors can accumulate, gradually changing the appearance of a previously visited scene. Second, we keep the per-chunk memory encoding cost independent of video length by retaining the latest frame and retrieving eight earlier frames. On particularly complex camera paths, the retrieved frames may not fully capture details visible from a narrow range of viewpoints.

\end{document}